\PassOptionsToPackage{table, dvipsnames}{xcolor}
\documentclass[10pt,twocolumn,letterpaper]{article}

\usepackage[pagenumbers]{cvpr} 

\usepackage[table]{xcolor}

\usepackage{booktabs}

\definecolor{cvprblue}{rgb}{0.21,0.49,0.74}
\usepackage[pagebackref,breaklinks,colorlinks,allcolors=cvprblue]{hyperref}
\usepackage{multirow}
\makeatletter
\newcommand{\rmnum}[1]{\romannumeral #1}
\newcommand{\Rmnum}[1]{\expandafter\@slowromancap\romannumeral #1@}
\usepackage{arydshln}
\newcommand{\thickhline}{
    \noalign {\ifnum 0=`}\fi \hrule height 1pt
    \futurelet \reserved@a \@xhline
}
\usepackage{bbding}
\usepackage{pifont}
\usepackage{wasysym}
\usepackage{amssymb}
\usepackage{comment}
\usepackage{makecell}

\definecolor{mygray}{gray}{0.93}
\definecolor{midblue}{HTML}{6ca6cd}
\definecolor{darkgreen}{HTML}{77A351}
\definecolor{darkblue}{HTML}{74B8EF}

\definecolor{mygreen}{RGB}{93,173,85}

\makeatother

\def\confName{CVPR}
\def\confYear{2025}

\usepackage{CJKutf8}

\newcommand{\worldwideweb}{\raisebox{-1.5pt}{\includegraphics[height=1.05em]{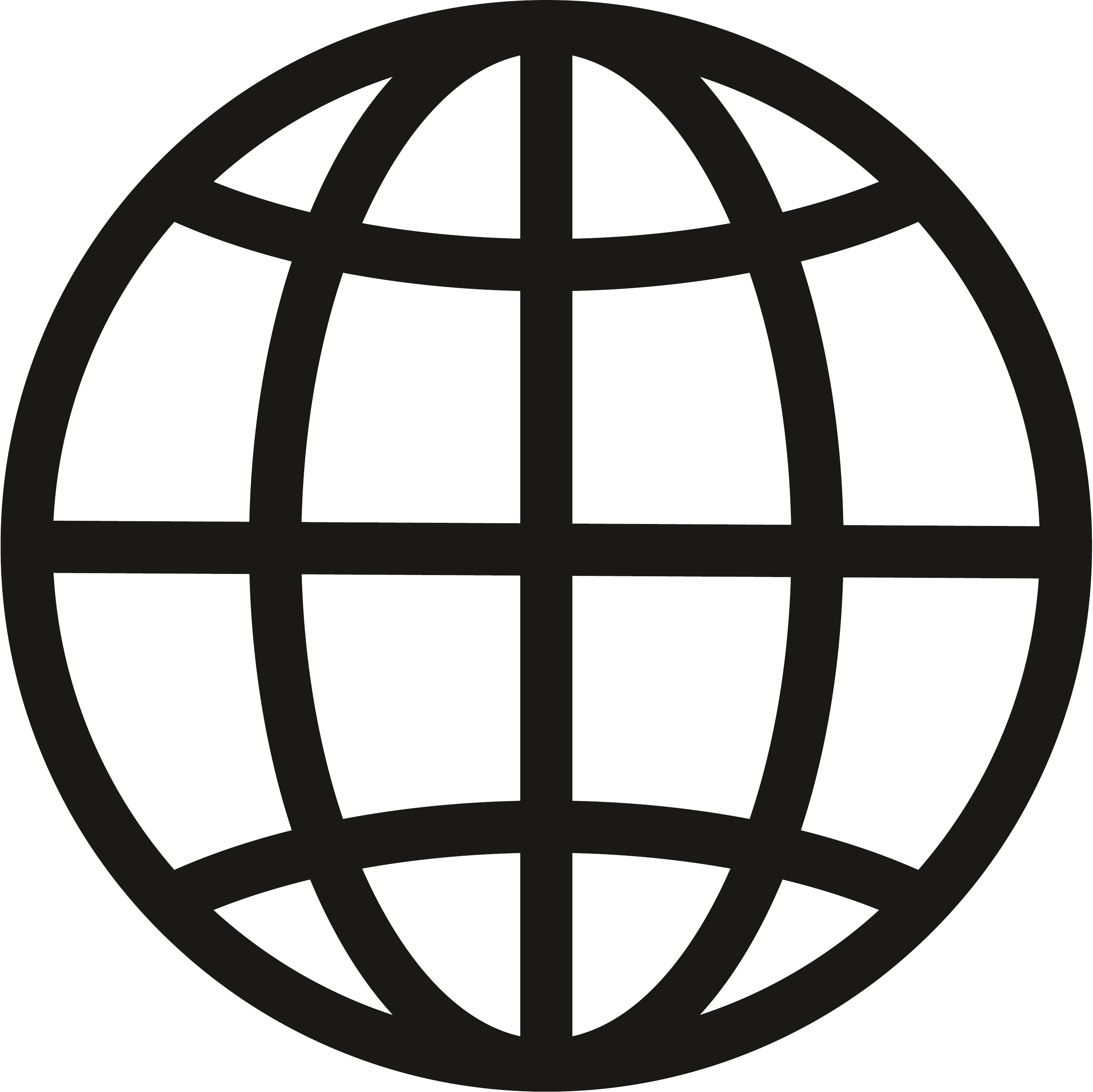}}\xspace}
\newcommand{\github}{\raisebox{-1.5pt}{\includegraphics[height=1.05em]{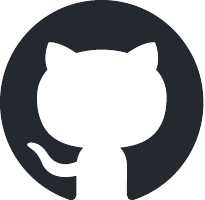}}\xspace}

\title{VideoRefer Suite: Advancing Spatial-Temporal Object Understanding\\ with Video LLM}

\author{
Yuqian Yuan$^{1,2}$\footnotemark[1], \ \ 
Hang Zhang$^2$, \ \ 
Wentong Li$^1$,  \ \  
Zesen Cheng$^2$, \ \  
Boqiang Zhang$^2$, \ \  
Long Li$^{1,2}$\footnotemark[1], \\  
Xin Li$^2$, \ \ 
Deli Zhao$^2$, \ \ 
Wenqiao Zhang$^1$\footnotemark[2], \ \
Yueting Zhuang$^1$, \ \ 
Jianke Zhu$^1$\footnotemark[2], \ \
Lidong Bing$^3$ \\[0.1cm]
$^1$Zhejiang University  \ \  
$^2$DAMO Academy, Alibaba Group \ \ 
$^3$Shanda AI Research Institute \\
\\
{\worldwideweb \href{https://damo-nlp-sg.github.io/VideoRefer/}{{\text{Project Page}}}} \quad \quad {\github \href{https://github.com/DAMO-NLP-SG/VideoRefer}{{\text{Code}}}}
}

\begin{document}
\maketitle

\renewcommand{\thefootnote}{\fnsymbol{footnote}}
\footnotetext[1]{Work is done during internship at DAMO Academy, Alibaba Group}
 \footnotetext[2]{Corresponding author}

\begin{abstract}
Video Large Language Models (Video LLMs) have recently exhibited remarkable capabilities in general video understanding.
However, they mainly focus on holistic comprehension and struggle with capturing fine-grained spatial and temporal details. 
Besides, the lack of high-quality object-level video instruction data and a comprehensive benchmark further hinders their advancements. 
To tackle these challenges, we introduce the VideoRefer Suite to empower Video LLM for finer-level spatial-temporal video understanding,
\textit{i.e.}, enabling perception and reasoning on any objects 
throughout the video.
Specially, we thoroughly develop VideoRefer Suite across three essential aspects: dataset, model, and benchmark. Firstly, we introduce a multi-agent data engine to meticulously curate a large-scale, high-quality object-level video instruction dataset, termed VideoRefer-700K. 
Next, we present the VideoRefer model, which equips a versatile spatial-temporal object encoder to capture precise regional and sequential representations.
Finally, we meticulously create a VideoRefer-Bench to comprehensively assess the spatial-temporal understanding capability of a Video LLM, evaluating it across various aspects.
Extensive experiments and analyses demonstrate that our VideoRefer model not only achieves promising performance on video referring
benchmarks but also facilitates general video understanding capabilities.

\end{abstract}

\section{Introduction}
\label{sec:intro}

\begin{figure*}[t]
  \centering
\includegraphics[width=0.999\linewidth]{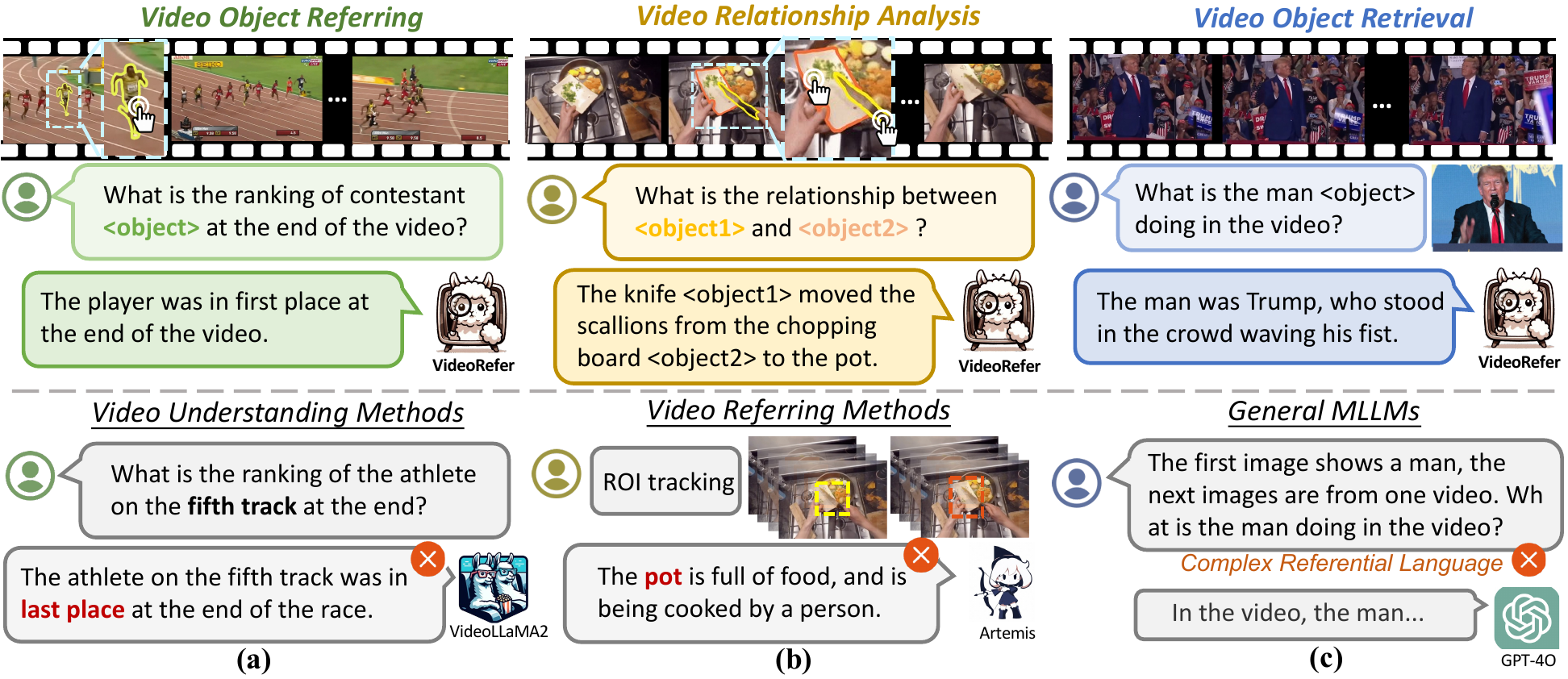}
\vspace{-5.0mm}
   \caption{Comparisons with previous general and specialized MLLMs. Our VideoRefer excels in multiple fine-grained regional and temporal video understanding tasks, including basic video object referring, complex video relationship analysis, and video object retrieval.}
   \label{fig:intro}
   \vspace{-2.0mm}
\end{figure*}

Multi-modal Large Language Models (MLLMs)~\cite{liu2023llava, liu2023improved, gpt4v,bai2023qwen-vl, reid2024gemini1_5, llavanext, lin2024vila, chen2024far} have 
demonstrated remarkable general-purpose capabilities for open-world image understanding through language-based dialogues over the past year.
In constant, extending their capabilities to the video domain presents unique challenges, as videos comprise dynamic sequences that not only showcase visual content but also convey the timing and relationships among various events and objects.
Currently, existing Video Large Language Models (Video LLMs)~\cite{lin2023video,cheng2024videollama, zhang2023video,maaz2023video,li2024mvbench,llavanext-video} primarily focus on holistic scene understanding. 
Unfortunately, these approaches often fall short in capturing the nuanced elements of video content. 
For instance, they often struggle to focus on 
user-specific objects, such as accurately describing a particular object.
Fig.~\ref{fig:intro}-(a) illustrates a typical example from general VideoLLaMA2~\cite{cheng2024videollama}.
The ability to discern such finer details in video content is crucial for applications that require 
precise object description, detailed event analysis, and predictive reasoning in dynamic environments.

To achieve fine-grained object understanding, numerous efforts have been devoted to image-based MLLMs, such as GPT4RoI~\cite{zhang2023gpt4roi}, Ferret~\cite{you2023ferret,zhang2024ferret} and  Osprey~\cite{yuan2024osprey}. 
These methods typically utilize a region encoder to obtain object-level embeddings, adapting them to LLMs for static image region-text alignment.
In contrast, research on video-based object understanding remains limited.
Some  works~\cite{wang2024elysium, yu2025merlin} directly convert the bounding box coordinates of object from specific frames into textual prompts to assist the LLM in identifying referred objects within the video.
However, these methods are plagued by impractical object referring and suffer from imprecise regional understanding. 
Alternatively, Artemis~\cite{qiu2024artemis} employs an external RoI tracker to capture an object across the video and extract box-level features for aligning with the LLM. 
However, as illustrated in Fig.~\ref{fig:intro}-(b), it primarily focuses on single-object referencing using coarse box-level representations, 
which restricts its capacity to handle complex tasks, such as analyzing relationships among multiple objects and performing intricate reasoning.
Therefore, developing an interactive Video LLM that facilitates a comprehensive understanding of  objects  within video 
represents a nontrivial research challenge.

In this work, we revisit the design of the Video LLM for finer-level video understanding. We contend that achieving this necessitates three essential components: a large-scale dataset containing high-quality object-level video instruction data, an architecture that integrates object embeddings with temporal cues, and a thorough benchmark for performance assessment.
To this end, we introduce \textit{VideoRefer Suite}, designed to empower Video LLMs with spatial-temporal object comprehension. 

\textbf{Dataset.} Firstly, to achieve regional alignment between video content and language embeddings, we meticulously curate a large-scale region-text video instruction dataset named \textbf{\texttt{VideoRefer-700K}}. 
Specifically, we present a multi-agent data engine to create high-quality video-based mask-text description pairs. This data engine leverages several expert models that excel in various tasks, collaborating meticulously to produce a diverse range of object-level instruction data for each object across the video. 
Our curated VideoRefer-700K comprises descriptions and multi-round QA pairs 
covering basic questions, complex reasoning and future predictions.

\textbf{Model.} Next, we introduce an effective Video LLM, named \textbf{\texttt{VideoRefer}}, that enables fine-grained perceiving, reasoning and retrieval for user-defined regions at any specified timestamps.
To accommodate both single-frame and multi-frame region inputs, we propose a versatile spatial-temporal object encoder. 
Specially, a Spatial Token Extractor is developed to generate accurate object-level encoding at any frame, leveraging a unified pixel-level mask representation to allow arbitrary free-form input regions.
We then propose an adaptive Temporal Token Merge Module, which captures temporal contextual information across multiple frames while producing flexible, enriched regional representations.
The image-level and object-level embeddings are interleaved with language instructions to form the input sequence for the LLM, facilitating a detailed object understanding of the input video.

\textbf{Benchmark.}  Furthermore, to evaluate the regional video understanding capabilities of a Video LLM comprehensively, we develop a benchmark named \textbf{\texttt{VideoRefer-Bench}}, which consists of two sub-benchmarks: \textbf{\texttt{VideoRefer-Bench$^{\texttt{D}}$}}, which focuses on description generation from four aspects, and \textbf{\texttt{VideoRefer-Bench$^{\texttt{Q}}$}}, which emphasizes multiple-choice question answering across five aspects.  VideoRefer-Bench thoroughly assesses the model's performance across various timestamps and objects, evaluating the abilities in comprehensive captioning and reasoning, complex multi-object relationships, and future predictions.

As illustrated in Fig.~\ref{fig:intro}, 
our VideoRefer unlocks a range of advanced finer-level video understanding capabilities, 
including basic video object referring, intricate relationship analysis among objects and object retrieval tasks, maintaining  user interactivity. 
In particular, VideoRefer can be seamlessly integrated with the off-the-shelf SAM 2~\cite{ravi2024sam} to further enhance user interactivity by enabling a comprehensive understanding of everything user click on.
Extensive experiments conducted on VideoRefer-Bench and general video understanding benchmarks, yield compelling results and demonstrate the efficacy of our approach.
Notably, VideoRefer not only significantly surpasses the state-of-the-art methods in regional video understanding across temporal, sequential and relationship reasoning, but also advances the general video understanding abilities.

\section{Preliminary}

\subsection{Background and Video-referring Task.}
To attain precise regional comprehension,  MLLMs can be incorporated with instance-level visual representations. This integration allows models to generate semantic understandings that focus on specific regions.
As for image-based MLLMs, recent researchs~\cite{zhang2023gpt4roi, yuan2024osprey,guo2024regiongpt,you2023ferret,chen2023shikra,chen2023position,zhang2024ferret,xuan2024pink, yue2024sc,zhao2023chatspot,rasheed2024glamm,cai2023making,tian2024chatterbox,zhan2024griffon,fei2024vitron}  has demonstrated a significant trend to enable the image referring with spatial visual prompts. In contrast, research focused on video-based regional understanding across sequential scenes is relatively limited.

The video referring task involves comprehending user-specific regions at designated moments or a time periods within a video~\cite{yu2025merlin,wang2024elysium,qiu2024artemis}. 
The basic video referring task focuses on captioning, 
while more complex tasks involve reasoning about the relationships between objects, and inferring their future states or interactions.
Video referring tasks can significantly enhance the functionality and applicability of video analysis for Video LLM across multiple domains, such as navigation, surveillance, and interactive robotics. 

\subsection{Task Formulation.}
For basic video object referring, the model processes questions phrased as ``What is \texttt{<object>} doing in this video?", where the \texttt{<object>} is specified by the user at a specific moment $t$ or over a duration of time. In more complex scenarios involving various object relationships, the model requires multiple user-defined regions, such as \texttt{<object1>}, \texttt{<object2>} and \texttt{<objectK>} along with the corresponding questions, like ``How do \texttt{<object1>} and \texttt{<object2>} interact with each other?''. 
To address these nuanced regional and temporal tasks, we provide a unified problem formulation. 

For a given video $V \in \mathbb{R}^{N \times W \times H \times C}$, where $N$, $W$, $H$, $C$ denote the frame number, height, width and channels, respectively. We define all the \texttt{<object>} as $\boldsymbol{R}$, where $\boldsymbol{R} = \{R_1, R_2, \ldots, R_n\}$. Here, $n$ represents the total number of objects  specified by the user.
 \(R_j\) is expressed as \(R_j = \{r_{ij} \mid i \in \boldsymbol{T}\}\), with $r_{ij}$ representing a region within a single frame, and \(\boldsymbol{T}\) being a set containing one or multiple timestamps.
For a Video LLM, 
the model optimization process aims to maximize the log-likelihood of generating text conditioned on $V$, $\boldsymbol{R}$, and text-based prompt $x$ across the entire training dataset to produce the desired output:
\begin{small}
\begin{equation}
    \mathcal{L} = \sum_{(V,\boldsymbol{R},x,y)} \log P(y \mid V, R_1,..., R_n, x),
\end{equation}
\end{small}
where $y$ denotes the ground truth label.

\section{VideoRefer Suite}
Our VideoRefer Suite consists of three crucial components: a comprehensive dataset, \textbf{\texttt{VideoRefer-700K}},
 containing high-quality instruction-following object-level annotations; 
 a Video LLM, \textbf{\texttt{VideoRefer}}, capable of pixel-level regional and temporal comprehension; and 
 an evaluation benchmark, \textbf{\texttt{VideoRefer-Bench}}, developed to evaluate models across various video referring tasks.

\begin{figure}[t]
  \centering
\includegraphics[width=0.999\linewidth]{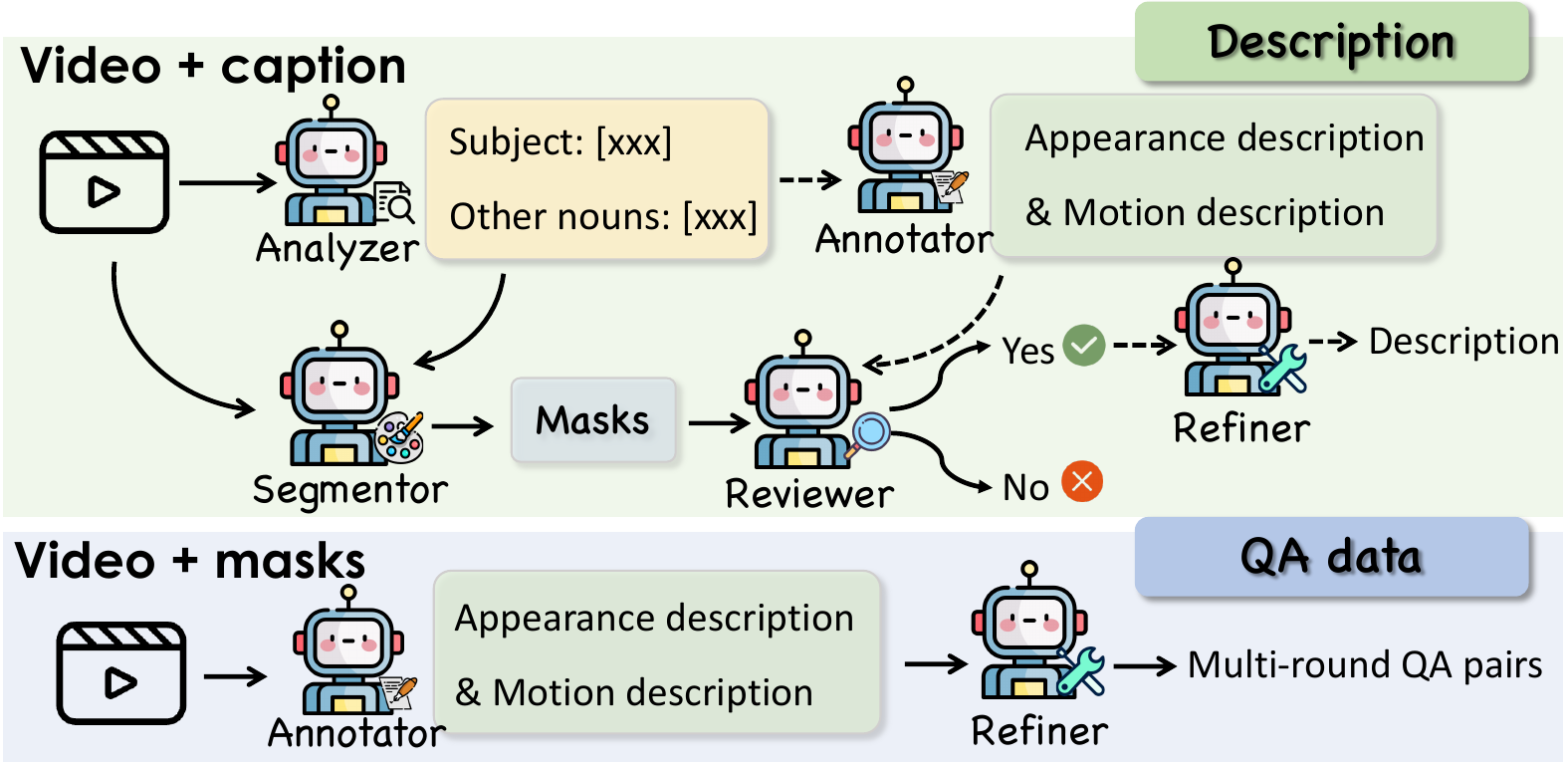}
\vspace{-5.0mm}
   \caption{A multi-agent data engine for the construction of our VideoRefer-700K.}
   \label{fig:agent-engine}
   \vspace{-4mm}
\end{figure}

\subsection{VideoRefer-700K Dataset}
\label{data_con}

\subsubsection{Multi-agent Data Engine}
We develop an automatic multi-agent data engine to create \textbf{\texttt{VideoRefer-700K}}, a \textit{large-scale} and \textit{high-quality} object-level video instruction-following dataset. Specially, we utilize off-the-shelf expert models that excel in tasks such as captioning, detection, segmentation and summation as collaborative agents to carefully create diverse types of object-level instruction data.
As illustrated in Fig.~\ref{fig:agent-engine},  our curation pipeline involves five components: (\rmnum{1}) Analyzer for noun extraction; (\rmnum{2}) Annotator for object-level caption generation; (\rmnum{3}) Segmentor for mask generation; (\rmnum{4}) Reviewer for correspondence verification; and (\rmnum{5}) Refiner for summarization\&refinement.
This multi-agent data engine effectively eliminates noisy or irrelevant contexts, ensuring that the data maintains its accuracy and relevance.

\textbf{Analyzer for Noun Extraction.}
Considering that most available video datasets contain the short scene-level caption, we begin by analyzing the raw captions to accurately capture the nouns within the sentences, \textit{i.e.}, objects occurred in the video scene.
To achieve this, we employ an Analyzer to extract nouns, encompassing both subjects and other relevant nouns. The Qwen2-Instruct-7B model ~\cite{yang2024qwen2} serves as our analytical tool in this process.

\textbf{Annotator for Object-level Caption Generation.}
To obtain detailed descriptions of the extracted nouns, we employ a general video understanding model as an annotator. We prompt the model to provide comprehensive descriptions focused specifically on the objects, rather than the holistic narrative of the whole video. 
To enhance accuracy and detail, we query the model twice: emphasizing dynamic actions\&movements, and highlighting static appearances\&states, respectively.  
Specifically, we filter out static actions related to the subjects to maintain variability and dynamism in the videos. The open-source InternVL2-26B model~\cite{chen2024far} serves as our annotator.

\textbf{Segmentor for Mask Generation.}
To acquire pixel-level masks as object-level region representations for each extracted noun, we first select a random frame from the video and extract the bounding box using Grounding-DINO~\cite{liu2023grounding} through open-set grounding, with the extracted noun serving as the input text prompt. 
Subsequently, HQ-SAM~\cite{ke2023segment} is employed to generate the high-quality mask based on the corresponding box prompt. To accommodate multi-frame input, we further generate masks for each video frame using SAM 2~\cite{ravi2024sam}.

\textbf{Reviewer for Correspondence Verification.}
To address potential errors and mismatches in this data construction pipeline, we introduce a Reviewer to verify the correspondence between masks and descriptions.  Initially, we employ Osprey~\cite{yuan2024osprey} to provide a region-level description for a specific frame. The Reviewer then assesses whether the descriptions from Osprey and the Annotator refer to the same object. After this filtering process, we retain only 40\% of samples to ensure accuracy. 
Qwen2-Instruct-7B model~\cite{yang2024qwen2} is chosen as the Reviewer for this task, due to its efficiency and suitability for handling the complexity of this process.

\textbf{Refiner for Summarization\&Refinement.}
Finally, we utilize a reliable Refiner, GPT-4o~\cite{gpt4o}, to summarize and refine the temporal and appearance-related captions generated by the annotator. This process aims to further eliminate repetition and hallucinations, ensuring a coherent and accurate final object-level instruction-following dataset.

\begin{figure*}[t]
  \centering
\includegraphics[width=0.99\linewidth]{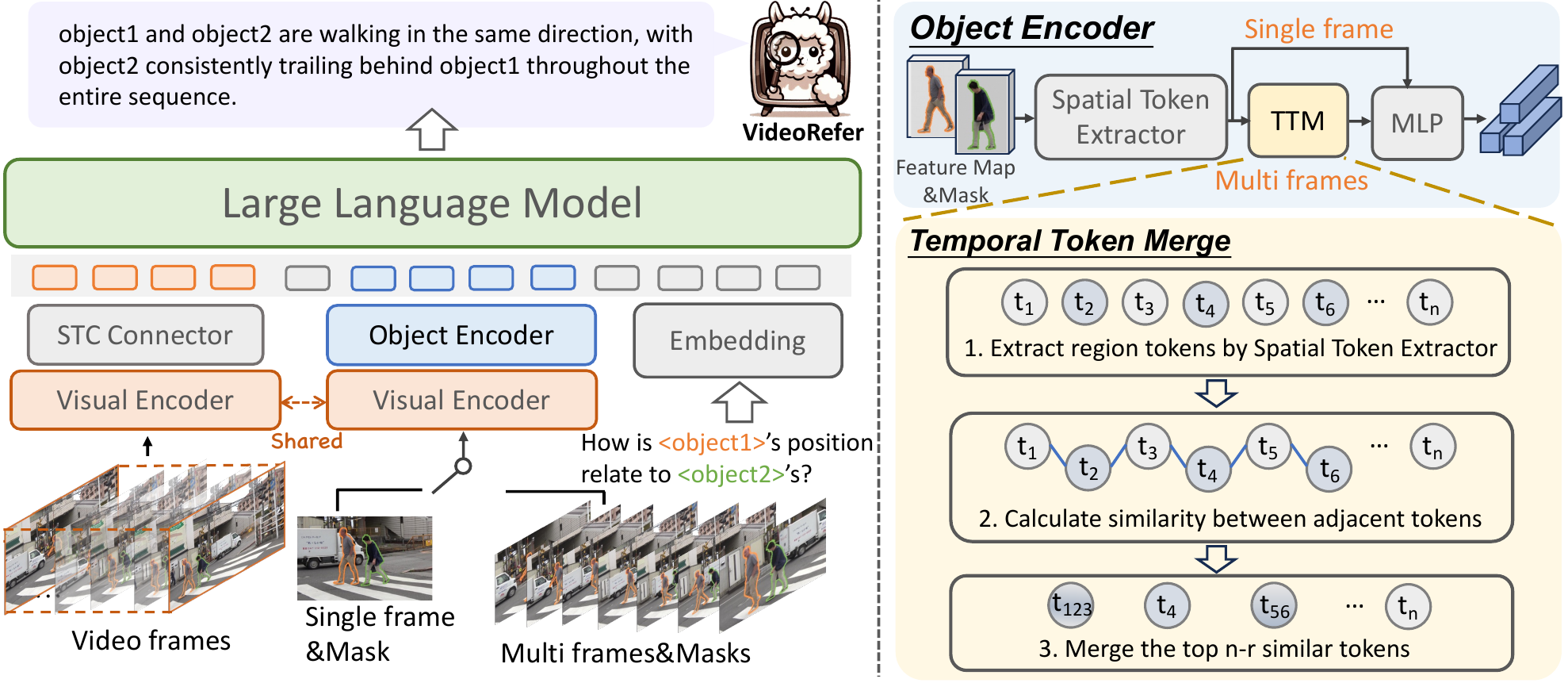}
\vspace{-1.5mm}
   \caption{Model architecture of our VideoRefer for spatial-temporal video object understanding.}
   \label{fig:method}
   \vspace{-5mm}
\end{figure*}

\subsubsection{Data Characteristics}

By leveraging our multi-agent data engine, we meticulously create three primary types of object-level video instruction data: detailed captions, short captions, and multi-round question-answer (QA) pairs.

\textbf{Object-level Detailed Caption.} We utilize a subset of large-scale Panda-70M~\cite{chen2024panda}, which has a short caption for each video. We generate 125K high-quality object-level detailed captions through our full multi-agent data engine.

\textbf{Object-level Short Caption.}
To generate short captions, primarily for aligning object-level encoder with the LLM for pre-training, we employ a portion of the pipeline, which only includes the Analyzer and Segmentor. Specifically, in the Analyzer, we extract only singular subject nouns, enabling the reusing of raw captions 
for short descriptions.
Using this approach, we produce 500K short captions.

\textbf{Object-level QA.}
To generate instruction data that explicitly specifies particular objects or their relationships, 
we collect MeViS~\cite{ding2023mevis}, Ref-YouTube-VOS~\cite{seo2020urvos} and A2D-Sentence datasets. Both provide reliable short descriptions with mask annotations for each object region. By utilizing these short descriptions and masked videos, we first employ Annotator to generate object-level descriptions for each region, and then employ Refiner to generate QA pairs related to the objects within the videos, using a variety of prompts. Three types of region-based QA data have been created: (\textit{\rmnum{1}}) \textbf{Basic Questions:} These cover object types, attributes, actions, locations, and interactions over time. (\textit{\rmnum{2}}) \textbf{Reasoning Questions:} These require reasoning and background knowledge to explain events without relying on specific visual details. (\textit{\rmnum{3}}) \textbf{Future Predictions:} These involve anticipating future actions or events related to a given object. We generate 75K QA pairs in total.

\subsection{VideoRefer Model}

\subsubsection{Overall Architecture}
In this section, we introduce the \textbf{\texttt{VideoRefer}} framework, which ensures the next token predictions of Video LLM, enabling fine-grained mask-level comprehension at any specific regions and any timestamps for a given video. Given that the current Video LLM already exhibits strong general scene-level video understanding capabilities, we develop our model upon a well-established Video LLM, VideoLLaMA2.1~\cite{cheng2024videollama}. Our primary innovation is to introduce a versatile and unified spatial-temporal object encoder to obtain object-level representations across video scenes.

The overall architecture of our framework is illustrated in Fig.~\ref{fig:method}.
VideoRefer adopts a visual encoder and STC connector~\cite{cheng2024videollama} to encode the global scene-level visual representations, a pretrained text tokenizer to capture the language embeddings, and an instruction-following LLM for language decoding. To achieve video referring, 
we present a versatile and unified spatial-temporal encoder, denoted as  $\mathbf{REnc}$,  to derive object-level representations.
For a specific object  $R_j\in \boldsymbol{R}$, we define $R_j = \{r_{ij} \mid i \in \boldsymbol{T}\}$, where each $r_{ij}$ represents a unified 2D binary mask $M$ designed to accommodate free-form input regions, assigning a value of $1$ inside the region and $0$ outside.
The set of objects \(\boldsymbol{R}\), along with the image feature map $\boldsymbol{Z}$ extracted from the shared visual encoder, is then fed into the introduced object encoder \(\mathbf{REnc}\), which generates enriched object-level tokens, expressed as \({\mathcal{T}}_{\boldsymbol{R}} = \mathbf{REnc}(\boldsymbol{R}, \boldsymbol{Z})\). Finally, the interleaved scene-level tokens $\mathcal{T}_{{\boldsymbol{Z}}}$, object-level tokens $\mathcal{T}_{\boldsymbol{R}}$ and linguistic tokens  $\mathcal{T}_{{\boldsymbol{x}}}$ are sent to the LLM to obtain the fine-grained semantic understandings $Y$, formulated as \({Y} = \mathbf{\Phi}({\mathcal{T}_{\boldsymbol{Z}}},{\mathcal{T}_{\boldsymbol{R}}},{\mathcal{T}_{\boldsymbol{x}}})\), where \(\mathbf{\Phi}\) denotes the LLM.

\subsubsection{A Versatile Spatial-Temporal Object Encoder}
\label{sec:object_enc}

To support various spatial-temporal video understanding tasks, our presented object encoder not only captures mask-level spatial features within the single frame at a specific timestamp, but also aggregates temporal information across multiple frames over a duration of time.
Consequently, we devise two modes for our  object encoder: single-frame and multi-frame.
For the sake of brevity for better illustration, we use a single object $R_j$ 
as an example.
If multiple objects are specified by the user, we adopt the same manner to extract features for each object individually.

\textbf{Single-Frame.}
For single-frame mode, the input consists of a randomly selected frame along with the corresponding regions specified by the user in that frame. Here, $\boldsymbol{T}$ contains only a randomly chosen timestamp.
To generate the object-level token representations, we present the \textbf{\textit{Spatial Token Extractor}}.
In detail,  the image feature is initially extracted by the shared visual encoder to generate the global image feature $\mathbf{F}_I \in \mathbb{R}^{1\times H_I\times W_I\times D_I}$, where $H_I$, $W_I$, $D_I$ denote the height, width and dimension of the image feature, respectively.
Each binary mask $M$ of an object is then resized to match the shape of the image feature.
We utilize the Mask Pooling operation upon image feature to extract object-level spatial feature ${\mathbf{F}_O}\in \mathbb{R}^{1\times D_I}$ for each mask, which pools all features within the region $M$ to generate an object-level representation. 
Finally, an MLP layer is employed to adapt and produce the object-level token ${\mathbf{O}}\in \mathbb{R}^{1\times C}$ for each object region.

\textbf{Multi-Frame.} 
In the multi-frame mode, the input consists of a list of selected frames from the video, accompanied with their respective object regions, \textit{i.e.}, $\boldsymbol{T}$ contains a list of timestamps from the video.
The frame-level feature is extracted using the shared visual encoder to generate the image feature $\mathbf{F}_I \in \mathbb{R}^{k \times H_I\times W_I\times D_I}$, where $k$ represents the number of selected frames. We then employ the Spatial Token Extractor to generate the object-level tokens for each frame. Hence, we obtain the object tokens ${\mathbf{O}}\in \mathbb{R}^{k\times C}$.
To aggregate distinct temporal object-level representations across multiple frames over a time duration while minimizing redundant tokens, we propose the \textit{\textbf{Temporal Token Merge Module}}, which is designed to effectively capture essential object-level tokens across the temporal dimension.
Specifically, starting with spatial object tokens ${\mathbf{O}}\in \mathbb{R}^{k\times C}$, we first compute the cosine similarity between each pair of adjacent tokens, formulated as:
\begin{equation}
\mathbf{S}_{m,m+1} = \frac{\mathbf{O}_m \cdot \mathbf{O}_{m+1}}{\| \mathbf{O}_m \| \cdot \| \mathbf{O}_{m+1} \|}, 0\leq m\textless k.
\end{equation}
Subsequently, we select the top $k-u$ similarity scores from $\mathbf{S}$, where $u$ is a predefined constant.  The corresponding pairs of tokens are then merged into a single union, resulting in $u$ unions.
For each union, we apply straightfoward average pooling to produce a single  distinct representative token.  Ultimately, $u$ tokens, represented as ${\mathbf{O}}\in \mathbb{R}^{u\times C}$, are generated following an MLP layer for each object, ensuring both spatial integrity and temporal coherence without disrupting spatial structure.

\subsection{VideoRefer-Bench}
\begin{figure}[t]
  \centering
\includegraphics[width=0.999\linewidth]{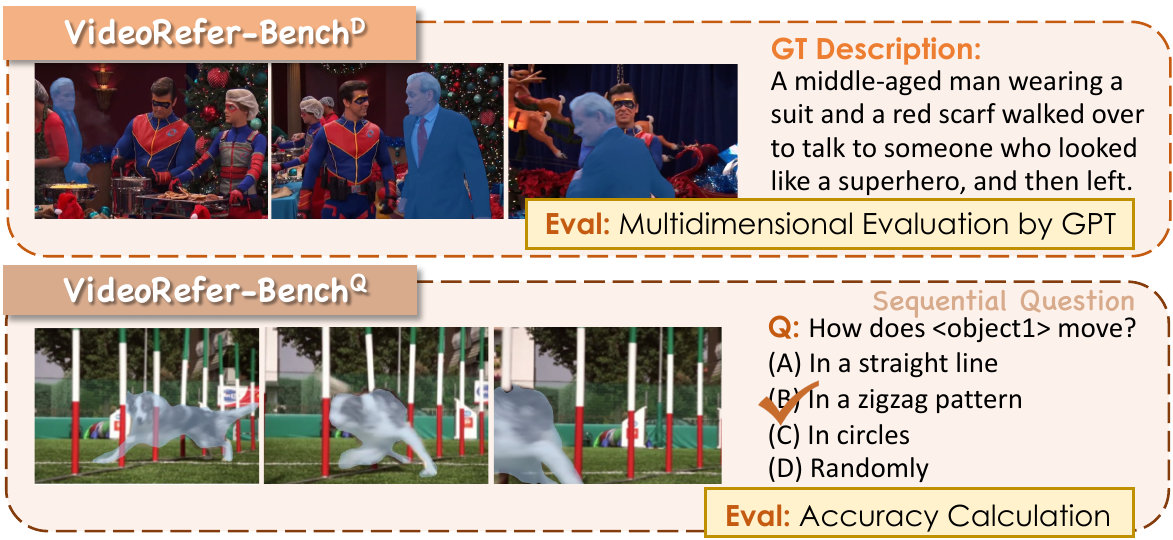}
\vspace{-5mm}
   \caption{Exemplar visual illustration of VideoRefer-Bench.}
   \label{fig:benchmark-1}
   \vspace{-5mm}
\end{figure}

\begin{figure}[t]
  \centering
\includegraphics[width=0.999\linewidth]{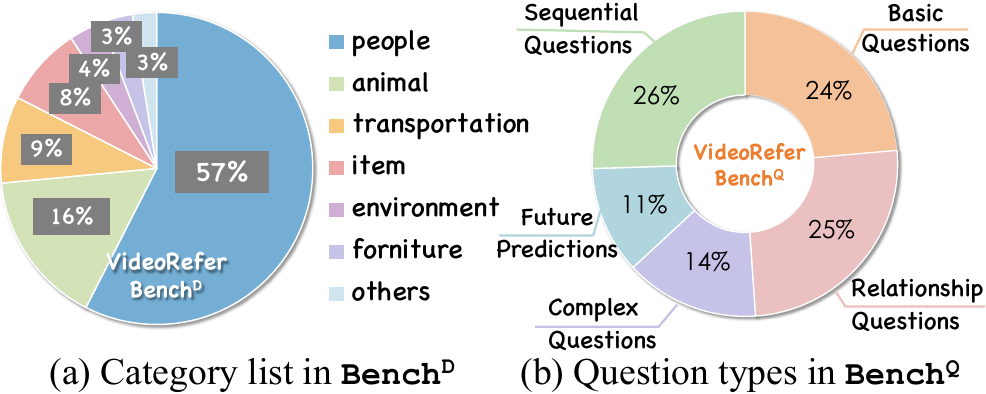}
\vspace{-4.5mm}
   \caption{Data characteristics of VideoRefer-Bench.}
   \label{fig:benchmark-2}
   \vspace{-3.5mm}
\end{figure}

To comprehensively evaluate the models' capability on video-based regional comprehension, we have developed a benchmark named \textbf{\texttt{VideoRefer-Bench}}. This benchmark assesses the models in two key areas: \textit{Description Generation}, corresponding to VideoRefer-Bench$^\text{D}$, and \textit{Multiple-choice Question-Answer}, corresponding to VideoRefer-Bench$^\text{Q}$. 
Fig.~\ref{fig:benchmark-1} and Fig.~\ref{fig:benchmark-2} provide exemplar visual illustrations and data characteristics, respectively.

\subsubsection{VideoRefer-Bench$^{\textbf{D}}$}
We introduce a sub-benchmark, \textbf{\texttt{VideoRefer-Bench$^{\texttt{D}}$}} 
specifically designed to evaluate the description generation performance of video-based referring models. The benchmark comprises a total of 400 curated data entries. We curated the test set based on Panda-70M~\cite{chen2024panda}, employing the pipeline described in Section~\ref{data_con}, followed by a meticulous human check. Furthermore, we developed an evaluation pipeline utilizing the GPT-4o model. This pipeline rigorously assesses various capabilities of the model by assigning scores to the generated predictions on a scale range from 0 to 5 across the following four dimensions:

\begin{itemize}
    \item \textbf{Subject Correspondence (SC):} This dimension evaluates whether the subject of the generated description accurately corresponds to that specified in the ground truth.
    
    \item \textbf{Appearance Description (AD):} This criterion assesses the accuracy of appearance-related details, including color, shape, texture, and other relevant visual attributes.
    
    \item \textbf{Temporal Description (TD):} This aspect analyzes whether the representation of the object's motion is consistent with the actual movements.
    
    \item \textbf{Hallucination Detection (HD):} This facet identifies discrepancies by determining if the generated description includes any facts, actions, or elements absent from reality, like imaginative interpretations or incorrect inferences.
\end{itemize}

\subsubsection{VideoRefer-Bench$^{\textbf{Q}}$}
The other sub-benchmark \textbf{\texttt{VideoRefer-Bench$^{\texttt{Q}}$}} is designed to evaluate 
the proficiency of MLLMs in interpreting video objects.
 We meticulously curated a dataset comprising 198 videos sourced from various datasets, including DAVIS-2017~\cite{pont20172017} and the test set of MeViS~\cite{ding2023mevis}. 
 To facilitate a robust evaluation, we annotated a set of 1,000 high-quality multiple-choice questions.  These questions are crafted to assess different dimensions of understanding, including \textbf{Basic Questions}, \textbf{Sequential Questions}, \textbf{Relationship Questions}, \textbf{Reasoning Questions}, and \textbf{Future Predictions}.
The annotations were performed by researchers with extensive research experience in vision-language learning. Importantly, each QA pair is required to be explicitly linked to a specific video region. This ensures that the MLLMs cannot provide answers without actually analyzing the video or the designated object.

\begin{table*}[t]
  \centering
  \resizebox{0.98\textwidth}{!}{
  \setlength{\tabcolsep}{4.0mm}{
  \begin{tabular}{lcccccccccc}
    \thickhline
    {\multirow{2}{*}{\textbf{Method}}} & \multicolumn{5}{c}{\textbf{Single-Frame}} & \multicolumn{5}{c}{\textbf{Multi-Frame}} \\
    \cmidrule(r){2-6}
    \cmidrule(r){7-11}
    & \textbf{SC} & \textbf{AD} & \textbf{TD} & \textbf{HD} & \textbf{Avg.} &\textbf{SC} & \textbf{AD} & \textbf{TD} & \textbf{HD} & \textbf{Avg.}\\
    \hline
    \hline
    \rowcolor{black!5} \multicolumn{11}{c}{\textit{Generalist Models}} \\
    \hline
    \textcolor{black!50}{LongVU-7B~\cite{shen2024longvu}} & \textcolor{black!50}{2.02} & \textcolor{black!50}{1.45} & \textcolor{black!50}{1.98} & \textcolor{black!50}{1.12} & \textcolor{black!50}{1.64} & \textcolor{black!50}{2.33} & \textcolor{black!50}{1.80} & \textcolor{black!50}{2.39} & \textcolor{black!50}{1.68} & \textcolor{black!50}{2.05} \\
    \textcolor{black!50}{LongVA-7B~\cite{zhang2024longva}} & \textcolor{black!50}{2.63} & \textcolor{black!50}{1.59} & \textcolor{black!50}{2.12} & \textcolor{black!50}{2.10} & \textcolor{black!50}{2.11} & \textcolor{black!50}{3.02} & \textcolor{black!50}{2.30} & \textcolor{black!50}{1.92} & \textcolor{black!50}{2.51} & \textcolor{black!50}{2.44} \\
    \textcolor{black!50}{LLaVA-OV-7B~\cite{li2024llava}} & \textcolor{black!50}{2.62} & \textcolor{black!50}{1.58} & \textcolor{black!50}{2.19} & \textcolor{black!50}{2.07} & \textcolor{black!50}{2.12} & \textcolor{black!50}{3.09} & \textcolor{black!50}{1.94} & \textcolor{black!50}{2.50} & \textcolor{black!50}{2.41} & \textcolor{black!50}{2.48} \\
    \textcolor{black!50}{Qwen2-VL-7B~\cite{yang2024qwen2}} & \textcolor{black!50}{2.97} & \textcolor{black!50}{2.24} & \textcolor{black!50}{2.03} & \textcolor{black!50}{2.31} & \textcolor{black!50}{2.39} & \textcolor{black!50}{3.30} & \textcolor{black!50}{2.54} & \textcolor{black!50}{2.22} & \textcolor{black!50}{2.12} & \textcolor{black!50}{2.55} \\
    \textcolor{black!50}{InternVL2-26B~\cite{chen2024far}} & \textcolor{black!50}{3.55}&\textcolor{black!50}{\underline{2.99}}&\textcolor{black!50}{2.57}&\textcolor{black!50}{2.25}&\textcolor{black!50}{2.84}&\textcolor{black!50}{4.08} & \textcolor{black!50}{3.35} & \textcolor{black!50}{3.08} & \textcolor{black!50}{2.28} & \textcolor{black!50}{3.20} \\
    \textcolor{black!50}{GPT-4o-mini~\cite{gpt4o}} & \textcolor{black!50}{\underline{3.56}}&\textcolor{black!50}{2.85}&\textcolor{black!50}{2.87}&\textcolor{black!50}{2.38}&\textcolor{black!50}{2.92}&\textcolor{black!50}{3.89} & \textcolor{black!50}{3.18} & \textcolor{black!50}{2.62} & \textcolor{black!50}{2.50} & \textcolor{black!50}{3.05} \\
    \textcolor{black!50}{GPT-4o~\cite{gpt4o}} &\textcolor{black!50}{3.34} &\textcolor{black!50}{2.96}&\textcolor{black!50}{\underline{3.01}}&\textcolor{black!50}{2.50}&\textcolor{black!50}{\underline{2.95}}& \textcolor{black!50}{\underline{4.15}} & \textcolor{black!50}{\textbf{3.31}} & \textcolor{black!50}{\textbf{3.11}} & \textcolor{black!50}{2.43} & \textcolor{black!50}{\underline{3.25}} \\
    \hline
    \rowcolor{black!5} \multicolumn{11}{c}{\textit{Specialist Models}} \\
    \hline
    \textbf{\textit{Image-level models}}&&&&& \\
    Ferret-7B~\cite{you2023ferret} &3.08&2.01&1.54&2.14&2.19 & 3.20 & 2.38 & 1.97 & 1.38& 2.23\\ 
    Osprey-7B~\cite{yuan2024osprey} &3.19 & 2.16 & 1.54 & 2.45 & 2.34 & 3.30 & 2.66 & 2.10 & 1.58 & 2.41\\
    \cdashline{1-11}[1pt/1pt]
    \textbf{\textit{Video-level models}}&&&&& \\
    Elysium-7B~\cite{wang2024elysium} & 2.35 & 0.30 & 0.02 & \textbf{3.59} & 1.57 & --& --&--&--&-- \\
    Artemis-7B~\cite{qiu2024artemis} & -- & -- & -- & -- & -- &3.42&1.34&1.39&\underline{2.90}&2.26 \\
    \rowcolor{blue!5} \textbf{VideoRefer-7B} & \textbf{4.41} & \textbf{3.27} & \textbf{3.03} & \underline{2.97} & \textbf{3.42} & \textbf{4.44} & \underline{3.27} & \underline{3.10} & \textbf{3.04} & \textbf{3.46} \\
    \hline
  \end{tabular}}}
  \vspace{-1.5mm}
  \caption{Performance comparisons on VideoRefer-Bench$^\text{D}$. The best results are \textbf{bold} and the second-best results are \underline{underlined}. ``--'' means that the model does not support the certain input form. \textcolor{black!50}{Grey entries} denote cases where the original method cannot accomplish the task; for these tests, masks of the targets were overlaid on the original video (the same below).}
  \vspace{-1.5mm}
  \label{tab:main1}
\end{table*}

\begin{table*}[t]
  \centering
  \resizebox{0.98\textwidth}{!}{
  \setlength{\tabcolsep}{4.5mm}{
  \begin{tabular}{l||c|c|c|c|c||c}
    \thickhline 
     & \textbf{Basic} & 
     \textbf{Sequential} &
    \textbf{Relationship} & 
    \textbf{Reasoning} & 
    \textbf{Future} & 
     \\
     {\multirow{-2}{*}{\textbf{Method}}} & \textbf{Questions} & \textbf{Questions} & \textbf{Questions} & \textbf{Questions} & \textbf{Predictions} & {\multirow{-2}{*}{\textbf{Average}}}\\
    \hline
    \hline
    \rowcolor{black!5} \multicolumn{7}{c}{\textit{Generalist Models}} \\
    \hline
    \textcolor{black!50}{LongVU-7B~\cite{shen2024longvu}} & \textcolor{black!50}{47.2} & \textcolor{black!50}{61.3} & \textcolor{black!50}{57.5} & \textcolor{black!50}{85.3} & \textcolor{black!50}{65.8} & \textcolor{black!50}{61.0} \\
    \textcolor{black!50}{LongVA-7B~\cite{zhang2024longva}} & \textcolor{black!50}{56.2} & \textcolor{black!50}{62.5} & \textcolor{black!50}{52.0} & \textcolor{black!50}{83.9} & \textcolor{black!50}{65.8} & \textcolor{black!50}{61.8} \\
    \textcolor{black!50}{InternVL2-26B~\cite{chen2024far}} & \textcolor{black!50}{58.5} & \textcolor{black!50}{63.5} & \textcolor{black!50}{53.4} & \textcolor{black!50}{\underline{88.0}} & \textcolor{black!50}{\textbf{78.9}} & \textcolor{black!50}{65.0} \\
    \textcolor{black!50}{GPT-4o-mini~\cite{gpt4o}}  & \textcolor{black!50}{57.6} & \textcolor{black!50}{67.1} & \textcolor{black!50}{56.5} & \textcolor{black!50}{85.9} & \textcolor{black!50}{75.4} & \textcolor{black!50}{65.8} \\
    \textcolor{black!50}{Qwen2-VL-7B~\cite{yang2024qwen2}}  & \textcolor{black!50}{62.0} & \textcolor{black!50}{\underline{69.6}} & \textcolor{black!50}{54.9} & \textcolor{black!50}{87.3} & \textcolor{black!50}{74.6} & \textcolor{black!50}{66.0} \\
    \textcolor{black!50}{LLaVA-OV-7B~\cite{li2024llava}} & \textcolor{black!50}{58.7} & \textcolor{black!50}{62.9} & \textcolor{black!50}{64.7} & \textcolor{black!50}{87.4} & \textcolor{black!50}{76.3} & \textcolor{black!50}{67.4}\\
    \textcolor{black!50}{GPT-4o~\cite{gpt4o}}       & \textcolor{black!50}{\underline{62.3}} & \textcolor{black!50}{\textbf{74.5}} & \textcolor{black!50}{\textbf{66.0}} & \textcolor{black!50}{\underline{88.0}} & \textcolor{black!50}{73.7} & \textcolor{black!50}{\underline{71.3}} \\
    \hline
    \rowcolor{black!5} \multicolumn{7}{c}{\textit{Specialist Models}} \\
    \hline
    Osprey-7B~\cite{yuan2024osprey} & 45.9 & 47.1 & 30.0 & 48.6 & 23.7 & 39.9\\
    Ferret-7B~\cite{you2023ferret} & 35.2 & 44.7 & 41.9 & 70.4 & 74.6 & 48.8 \\ 
    \cdashline{1-7}[1pt/1pt]
    \rowcolor{blue!5} \textbf{VideoRefer-7B} & \textbf{75.4} & 68.6 & \underline{59.3} & \textbf{89.4} & \underline{78.1} & \textbf{71.9} \\
    \hline
  \end{tabular}}}
  \vspace{-2.0mm}
  \caption{Performance comparisons on VideoRefer-Bench$^\text{Q}$. \textbf{Note}:
  Video-level specialist models, including Elysium~\cite{wang2024elysium} and Artemis~\cite{qiu2024artemis}, do not have the ability to handle multi-choice questions on VideoRefer-Bench$^\text{Q}$. }
  \vspace{-4.5mm}
  \label{tab:main2}
\end{table*}

\section{Experiments}

\subsection{Implementation Details}
We adopt \texttt{siglip-so400m-patch14-384}~\cite{zhai2023sigmoid} as the vision encoder, Qwen-2~\cite{yang2024qwen2} as the LLM.
The AdamW~\cite{loshchilov2016sgdr} is used as the optimizer and the cosine annealing scheduler~\cite{loshchilov2017decoupled} is used to adjust learning rate.
We use a hybrid strategy including both single-frame and multi-frame modes during training.
We leverage a progressive training scheme, which consists of image-text alignment pre-training (Stage 1), region-text alignment pre-training (Stage 2), high-quality knowledge learning (Stage 2.5) and visual instruction tuning (Stage 3) stages, respectively. 
\textit{Please refer to the Appendix for detailed introduction to each stage}. At the first and second stages, we set global batch size to 256 and learning rate to 1$\times$10$^{-3}$ for one epoch. In stage 2.5 and stage 3, the learning rate is reduced to 2$\times$10$^{-5}$ with a global batch size of 128 for one epoch. Unless otherwise specified, all models adopt the 7B LLM.

\subsection{Main Results}
To evaluate the efficacy of our VideoRefer model, we conduct experiments on both
video referring tasks and general video understanding tasks to demonstrate its capabilities.

\subsubsection{Video Referring Tasks}

\begin{figure*}[t]
  \centering
\includegraphics[width=0.97\linewidth]{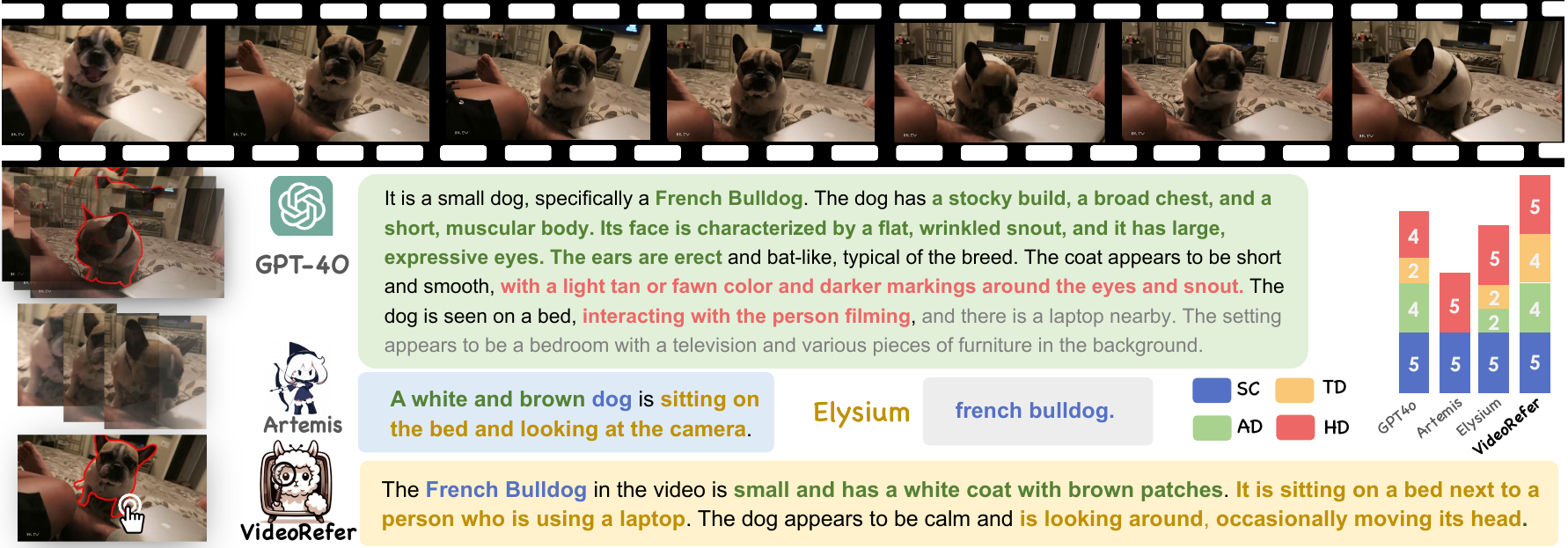}
\vspace{-1.0mm}
   \caption{Visual comparisons between our VideoRefer with general GPT-4o and regional video-level Elysium and  Artemis. Here we provide detailed illustrations on VideoRefer-Bench$^\text{D}$.}
   \label{fig:comp}
   \vspace{-3.0mm}
\end{figure*}

\textbf{VideoRefer-Bench$^{\textbf{D}}$.} 
We compare our approach on VideoRefer-Bench$^\text{D}$ with the previous generalist models, including GPT-4o~\cite{gpt4o}, GPT-4o-mini~\cite{gpt4o}, InternVL2~\cite{chen2024far}, Qwen2-VL~\cite{yang2024qwen2}, LLaVA-OV~\cite{li2024llava}, LongVA~\cite{zhang2024longva},  LongVU~\cite{shen2024longvu} and specialist models for object-level understanding, including image-level Osprey~\cite{yuan2024osprey}, Ferret~\cite{you2023ferret}, and video-level Elysium~\cite{wang2024elysium}, Artemis~\cite{qiu2024artemis}. 
Both single-frame and multi-frame modes are adopted for evaluation.  
In the single-frame mode, we select the first frame that contains the specific object with its aligned boundary for the generalist models. 
For image-level region understanding models, we utilize a random frame along with the corresponding region prompt as input.
In the multi-frame mode, we uniformly sample 16 frames with mask contours for generalist models. For image-level methods, we obtain the description frame by frame and then generate a summary using GPT-4o. For Elysium~\cite{wang2024elysium} and Artemis~\cite{qiu2024artemis}, we adhere to the official settings provided in their respective papers.
For our VideoRefer, we randomly select a single frame and uniformly sample 16 frames as inputs for the single-frame and multi-frame modes, respectively.
Table~\ref{tab:main1} presents the comparison results. Our approach achieves the leading average performance in regional-temporal video understanding compared to previous methods in both single-frame and multi-frame modes. 
Notably, VideoRefer attains top scores of 4.41, 3.27, and 3.03 for Subject Correspondence (SC), Appearance Description (AD), and Temporal Description (TD) in single-frame mode, and scores of 4.44 and 3.04 for SC and Hallucination Detection (HD) in multi-frame mode.
Fig.~\ref{fig:comp} illustrates a typical visual comparison.

\textbf{VideoRefer-Bench$^{\textbf{Q}}$.} We then compare our VideoRefer against the previous methods on VideoRefer-Bench$^\text{Q}$. 
Here, we set single-frame mode following settings of VideoRefer-Bench$^\text{D}$.
As shown in Table~\ref{tab:main2}, our VideoRefer achieves the best average performance with 71.9, which significantly outperforms the previous regional methods.  Especially, our approach excels in basic questions, relationship questions, reasoning questions and future predictions with 75.4, 59.3, 89.4 and 78.1 scores with best or second-best places, respectively. These results clearly demonstrate the superiority of our method in spatial-temporal video understanding.

\textbf{Previous Video Referring Metrics.} Following the previous state-of-the-art video referring approach, Artmis~\cite{qiu2024artemis}, we further conduct experiments on the test set of HC-STVG~\cite{tang2021human}.

Table~\ref{tab:hc-stvg} presents the comparison results. 
Our approach outperforms Artmis~\cite{qiu2024artemis} by \textcolor{mygreen}{+1.0}\%, \textcolor{mygreen}{+0.7}\%, \textcolor{mygreen}{+1.6}\%, \textcolor{mygreen}{+15.4}\%, and \textcolor{mygreen}{+2.9}\% on BLEU\@4~\cite{papineni2002bleu}, METEOR~\cite{banerjee2005meteor}, ROUGE\_L~\cite{lin2004rouge}, CIDEr~\cite{vedantam2015cider} and SPICE~\cite{anderson2016spice} metrics. These results demonstrate the superiority of our VideoRefer.

\vspace{-2mm}
\definecolor{mygreen}{RGB}{0, 128, 0}
\subsubsection{General Video Understanding}
To demonstrate the capabilities of our method, we conduct performance evaluation on general video understanding tasks. 
As shown in Table~\ref{tab:generalvideounder}, VideoLLaMA2.1~\cite{cheng2024videollama} achieves scores of 54.9\% on  Perception-Test~\cite{patraucean2024perception}, 57.3\% on MVBench~\cite{li2024mvbench}, and 54.9\%/56.4\% on VideoMME~\cite{fu2024video}. Based on that, our VideoRefer exhibits performance gains of \textcolor{mygreen}{+1.4}\%, \textcolor{mygreen}{+2.3}\%, and \textcolor{mygreen}{+1.0}\%/\textcolor{mygreen}{+1.2}\%, respectively. In contrast, Artemis demonstrates subpar performance. 
These results clearly indicate that our approach not only excels in object-level analysis, but also 
enhances the ability of general video understanding.

\begin{table}[t]
  \centering
  \renewcommand\arraystretch{1.17}
  \resizebox{0.48\textwidth}{!}{
  \setlength{\tabcolsep}{1.0mm}{
  \begin{tabular}{l|cccccc}
    \thickhline 
        \textbf{Method} &  \textbf{BLEU@4} & \textbf{METEOR}  &\textbf{ROUGE\_L} & \textbf{CIDER} & \textbf{SPICE}  \\
    \hline\hline
     Merlin~\cite{yu2025merlin} &3.3&11.3&26.0 & 10.5&20.1  \\
     Artemis~\cite{qiu2024artemis} & 15.5&18.0&40.8 & 53.2&25.4 \\
      \cdashline{1-4}[1pt/1pt]
     \rowcolor{blue!5}
     \textbf{VideoRefer} & \textbf{16.5} & \textbf{18.7} &\textbf{42.4}& \textbf{68.6} & \textbf{28.3}  \\
    \hline
    \end{tabular}}}
  \vspace{-2.5mm}
  \caption{Exprimental results on video-based referring metrics on the HC-STVG~\cite{tang2021human} test set.}
  \vspace{-2.5mm}
  \label{tab:hc-stvg}
\end{table}

\begin{table}[t]
  \centering
  \renewcommand\arraystretch{1.17}
  \resizebox{0.48\textwidth}{!}{
  \setlength{\tabcolsep}{1.0mm}{
  \begin{tabular}{l|cccc}
    \thickhline 
        \textbf{Method} &  \textbf{Perception-Test} & \textbf{MVBench}  & \textbf{VideoMME}  \\
    \hline\hline
     VideoLLaMA2~\cite{cheng2024videollama} & 51.4  & 54.6 & 47.9/50.3  \\
     VideoLLaMA2.1~\cite{cheng2024videollama} & 54.9  & 57.3 & 54.9/56.4   \\
      \cdashline{1-4}[1pt/1pt]
      Artemis~\cite{qiu2024artemis} &47.1 & 34.1&28.8/35.3 \\
     \rowcolor{blue!5}
     \textbf{VideoRefer} & \textbf{56.3} & \textbf{59.6} & \textbf{55.9/57.6}  \\
    \hline
    \end{tabular}}}
  \vspace{-2.0mm}
  \caption{Exprimental results on general video understanding tasks.}
  \vspace{-4.5mm}
  \label{tab:generalvideounder}
\end{table}

\subsection{Ablation Study}

\textbf{Single-frame \textit{vs.} Multi-frame.} We first validate the impacts on the single-frame and multi-frame modes, \textit{i.e.} with or without Temporal Token Merge (TTM) module to encode the multi-frame sequences during the inference.
As shown in Table~\ref{tab:frame_ablation}, our approach utilizing multi-frame mode exhibits improvements over the single-frame mode in both VideoRefer-Bench$^\text{D}$ and VideoRefer-Bench$^\text{Q}$ across all metrics. Notably, for sequential relation-based metrics, including Temporal Description (TD), Sequential Questions (SQ), and Relationship Questions (RQ), as well as hallucination-related metrics such as Hallucination Detection (HD), multi-frame mode showcases the superiority.

\textbf{Ablation on VideoRefer-700K Dataset.} 
Table \ref{tab:data_ablation} summarizes the ablation results for various data types in the constructed VideoRefer-700K dataset. The results indicate that using a short description yields a score of 2.43 on Bench$^\text{D}$ and 68.3 on Bench$^\text{Q}$, along with an MVBench score of 58.0. Incorporating question-answering (QA) data improves the performance to 2.45 for Bench$^\text{D}$ and 71.7 for Bench$^\text{Q}$, while maintaining an MVBench score of 58.4.  Notably, the method employing detailed descriptions achieves the best results, with scores of 3.42 on Bench$^\text{D}$, 71.9 on Bench$^\text{Q}$, and 59.6 on MVBench. These results demonstrate that the inclusion of more comprehensive data significantly enhances overall performance.

\begin{table}[t]
\centering
\renewcommand\arraystretch{1.17}
\scalebox{0.94}{
\setlength{\tabcolsep}{1.6mm}{
\begin{tabular}{l|ccc|ccc}
\thickhline 
\multirow{2}{*}{\textbf{Mode}} & \multicolumn{3}{c}{\textbf{VideoRefer-Bench$^\text{D}$}} & \multicolumn{3}{c}{\textbf{VideoRefer-Bench$^\text{Q}$}}  \\
\cmidrule(r){2-4} \cmidrule(r){5-7} 
 & TD & HD & Avg. & SQ & RQ & Avg.  \\
\hline
\hline
 Single-frame & 3.03 & 2.97  & 3.42 & 68.3 & 59.1  & 71.9  \\
\rowcolor{blue!5}  Multi-frame & \textbf{3.10} & \textbf{3.04} & \textbf{3.46} & \textbf{70.6}  & \textbf{60.5} & \textbf{72.1} \\
\hline
\end{tabular}
}}
\vspace{-2.5mm}
\caption{Results using different modes during the inference. Here, SQ and RQ are Sequential Questions and Relationship Questions. }
\vspace{-4.0mm}
\label{tab:frame_ablation}
\end{table}

\begin{table}[t]
\centering
\renewcommand\arraystretch{1.17}
\scalebox{0.92}{
\setlength{\tabcolsep}{1.7mm}{
\begin{tabular}{ll|cc|c}
\thickhline 
\multicolumn{2}{l|}{\textbf{Method}} & \textbf{Bench$^\text{D}$} & \textbf{Bench$^\text{Q}$} &\textbf{MVBench}  \\
\hline\hline
\rowcolor{black!5} 0 & w/o Regional data  & --  & --  & 57.9\\
 1 & \textbf{\textcolor{darkgreen}{+}} Short description & 2.43 & 68.3 & 58.0\\
 2 & \textbf{\textcolor{darkgreen}{+}} QA  & 2.45  & 71.7 &  58.4\\
\rowcolor{blue!5} 3 & \textbf{\textcolor{darkgreen}{+}}  Detailed description  & \textbf{3.42} & \textbf{71.9} & \textbf{59.6} \\
\hline
\end{tabular}
}}
\vspace{-2.5mm}
\caption{Ablation results on various data types in  VideoRefer-700K dataset. Bench denotes VideoRefer-Bench for simplicity.}
\vspace{-3.5mm}
\label{tab:data_ablation}
\end{table}

\textbf{Impacts of Different Union Numbers in TTM.}  
The Temporal Token Merge (TTM) Module is designed to capture essential object-level tokens across the temporal dimension in multi-frame mode.
Fig.~\ref{fig:tokenmerge} visualizes the  similarity scores between adjacent object token pairs. It is evident that most adjacent tokens exhibit high similarity, making it necessary to merge those tokens with significant similarity.
We conducted ablation experiments using temporal and sequential metrics to investigate the effects of varying numbers of token unions $u$.
The experimental results are detailed in Table~\ref{tab:token_num_ablation}. 
Notably, with $u=4$, VideoRefer achieves the best performance in Hallucination Detection (HD) and Sequential Questons (SQ), and ranks second in Reasoning Questions (RQ). We adopt $u=4$ to strike a balance between  performance and token costs in our approach.

\begin{table}[t]
\centering
\renewcommand\arraystretch{1.17}
\scalebox{0.94}{
\setlength{\tabcolsep}{2.5mm}{
\begin{tabular}{c|cc|cc}
\thickhline 
\multirow{2}{*}{\textbf{\makecell{Union\\$u$}}} & \multicolumn{2}{c}{\textbf{VideoRefer-Bench$^\text{D}$}} & \multicolumn{2}{c}{\textbf{VideoRefer-Bench$^\text{Q}$}}  \\
\cmidrule(r){2-3} \cmidrule(r){4-5} 
 & TD & HD  & SQ & RQ  \\
\hline
\hline
 32 & 3.17 & 3.01   & 68.7 & 58.1    \\
16 & \textbf{3.20} & 2.99 & 69.3 & 58.5   \\
8  & 3.18 & 3.02  & 69.6  & 57.8  \\
\rowcolor{blue!5} 4 & 3.10 & \textbf{3.04} & \textbf{70.6} & 60.5  \\ 
1 & 3.08 & 2.98  & 68.9 & \textbf{60.9}  \\
\hline
\end{tabular}
}}
\vspace{-1.mm}
\caption{Temporal and sequential performance comparisons for  various union $u$ in the TTM module under multi-frame mode. }
\vspace{-4.2mm}
\label{tab:token_num_ablation}
\end{table}

\section{Related Works}
\label{sec:relatedwork}
\subsection{Video Large Language Models}
Large Language Models (LLMs) have revolutionized the field of artificial intelligence by proving their capability to tackle diverse tasks related to language comprehension and generation. To fully leverage the potential of LLMs for visual understanding, researchers have increasingly turned their attention to image-based Multimodal Large Language Models (MLLMs)~\cite{liu2023llava, liu2023improved, gpt4v,bai2023qwen-vl, llavanext,lin2024vila,li2024tokenpacker, chen2024far,zhang2024hyperllava}, which integrate language and visual data within a unified feature space. This integration has emerged as a significant area of research focus.
In parallel, Video Large Language Models (Video LLMs)~\cite{cheng2024videollama,lin2023video,zhang2023video,maaz2023video,llavanext-video} have garnered increasing attention fueled by advancements in image-based MLLMs.
Most Video LLMs primarily follow the trend of utilizing pre-trained vision models to extract sequence-based information from videos, which is then interleaved with textual embeddings for LLM to generate responses~\cite{tang2023video}. 
Despite their promising results, current Video LLMs still face challenges in fine-grained regional and temporal understanding.

\subsection{Regional Understanding with MLLMs}
To attain fine-grained regional object-level comprehension,  MLLMs can be incorporated with  instance-level visual representations. This integration allows models to generate semantic understandings that focus on specific regions.
In the context of  image-based MLLMs, recent researchs~\cite{zhang2023gpt4roi, yuan2024osprey,guo2024regiongpt,you2023ferret,chen2023shikra,chen2023position,zhang2024ferret,xuan2024pink, yue2024sc,zhao2023chatspot,rasheed2024glamm,cai2023making,tian2024chatterbox,zhan2024griffon,fei2024vitron} has demonstrated a significant trend to enable the image referring with spatial visual prompts. 
In contrast, research focused on video-based regional understanding across dynamic sequence-based scenes is relatively limited.
Merlin~\cite{yu2025merlin} first explored video-based referring and future reasoning by employing three manually selected frames as visual input, which limits the model's ability to comprehend longer and more intricate scenes. Elysium~\cite{wang2024elysium} introduces a million-scale dataset for object-level tasks in videos; however, the provided descriptions tend to be quite simplistic.
Another reseach work is Artemis~\cite{qiu2024artemis}, but it primarily emphasizes basic single object descriptions, thereby constraining its capacity to analyze relationships among various objects or perform more complex tasks on specific objects within dynamic sequences. Moreover, Artemis utilizes a sparse bounding box representation, which inadequately captures the nuances of the objects.
Compounding these challenges is the lack of large-scale, high-quality region-level video instruction data and benchmarks for thorough evaluation, which further hampers progress in this domain. To address these issues, we introduce the VideoRefer Suite to advance spatial-temporal understanding.

\section{Conclusion}
In this work, we introduced the VideoRefer Suite to empower Video LLM for fine-grained spatial and regional video understanding.   Three key components have been proposed: 1) VideoRefer-700K: A large-scale, high-quality region-level video instruction data curated by a developed multi-agent engine; 2) VideoRefer: A Video LLM equipped with a versatile spatial-temporal object encoder that includes a Spatial Token Extractor and an adaptive Temporal Token Merge Module to enabling precise sequential regional representation; and 3) VideoRefer-Bench: a comprehensive benchmark that thoroughly evaluates model performance across multiple aspects, ensuring a holistic assessment of spatial-temporal capabilities.  Extensive experimental results and analyses have demonstrated the efficacy of our VideoRefer Suite, 
substantially advancing finer-level video understanding and analysis.

{
    \small
    \bibliographystyle{ieeenat_fullname}
    \bibliography{main}
}

\newpage
\appendix
\section*{Appendix}

\noindent

\begin{figure*}[t]
  \centering
\includegraphics[width=0.999\linewidth]{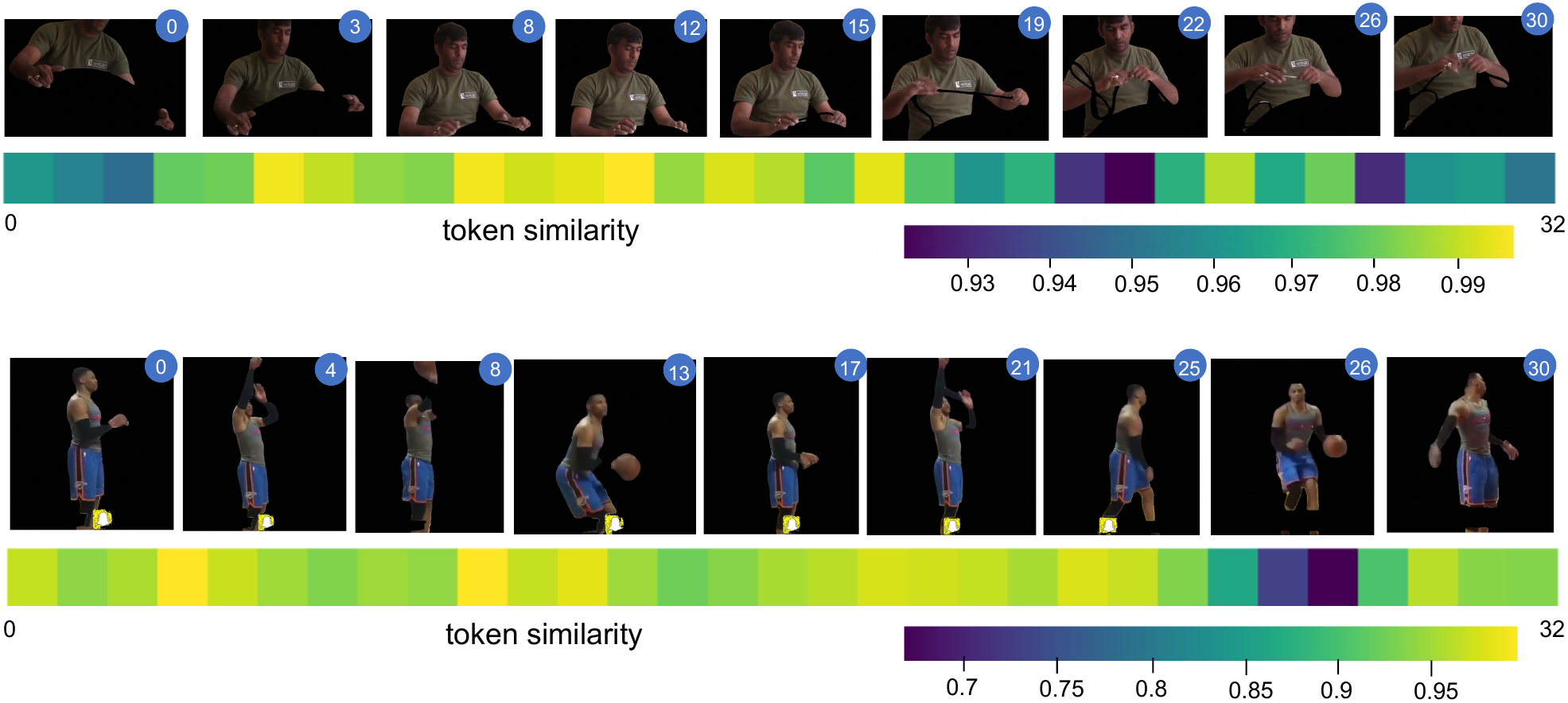}
\vspace{-1.0mm}
   \caption{Visualizations of similarity among adjacent object-level token pairs across the temporal dimension. Here, we use cosine similarity as the measurement.}
   \label{fig:tokenmerge}
\end{figure*}

\section{More Qualitative Results}

\begin{figure*}[t]
  \centering
\includegraphics[width=0.999\linewidth]{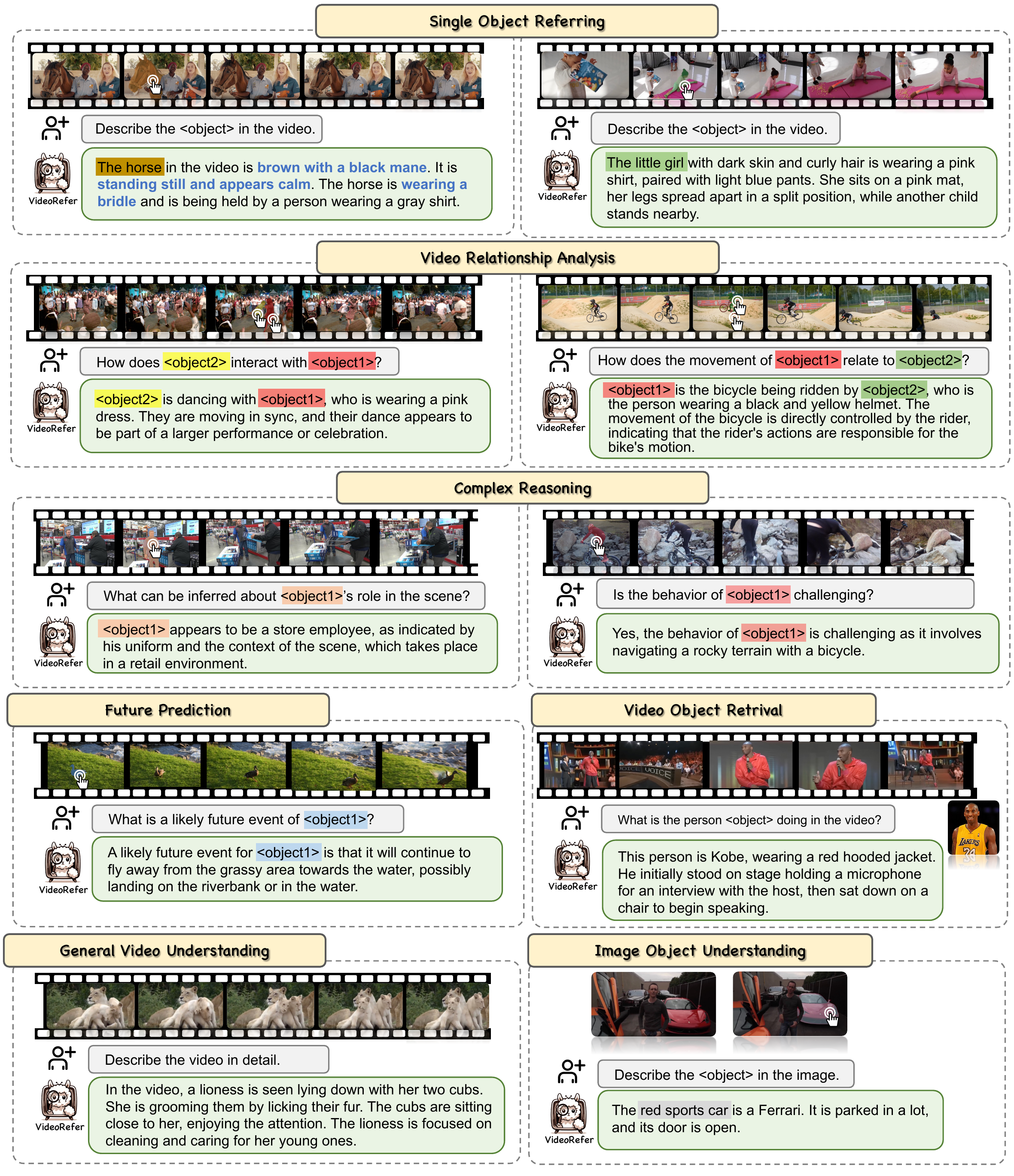}
\vspace{-4.5mm}
   \caption{Visualization results of VideoRefer across various tasks, including single-object referring, video relationship analysis, complex reasoning, future prediction, video object retrieval, as well as general video understanding and image object understanding.}
   \label{fig:vis}
\end{figure*}

We provide additional visualization results to emphasize performance across a variety of tasks, such as single-object referring, video relationship analysis, complex reasoning, future prediction, and video object retrieval. Besides, we present the examplar cases to demonstrate the capabilities in general video understanding and image object understanding.  
Fig.~\ref{fig:vis} showcases these visual examples. A randomly selected mask along with its corresponding frame is used as the region input.

\section{Additional Implemental Details}
\label{sec:implmetal}

\subsection{Training Stages}
\begin{figure}[t]
\centering
\includegraphics[width=0.98\linewidth]{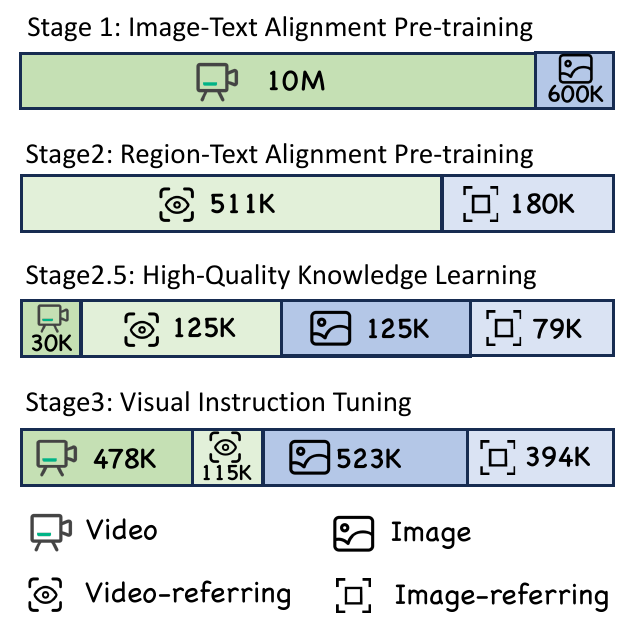}
\vspace{-2mm}
   \caption{Visual illustrations of the data distribution for each training stage.}
   \label{fig:train_data}
   \vspace{-3.5mm}
\end{figure}

The training pipeline of our model is structured into four distinct stages. Fig.~\ref{fig:train_data} presents the data distribution for each stage.

\begin{figure}[t]
  \centering
\includegraphics[width=0.8\linewidth]{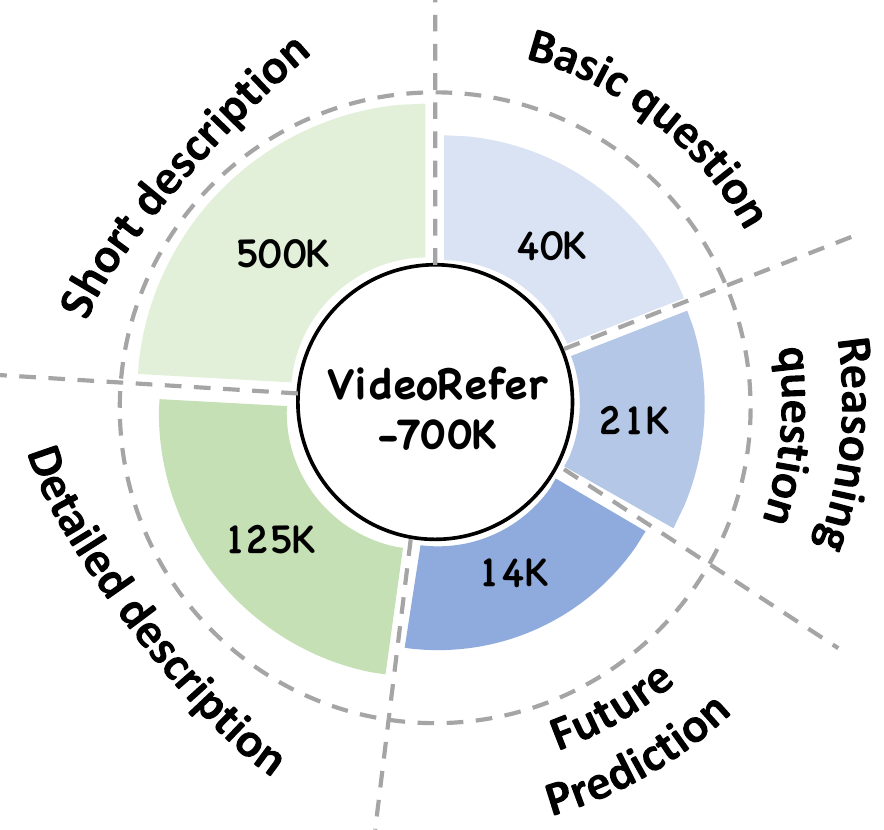}
   \caption{Data distributions of our VideoRefer-700K dataset, encompassing five different data types.}
   \label{fig:videorefer_data}
\end{figure}

\begin{figure*}[t]
  \centering
\includegraphics[width=0.93\linewidth]{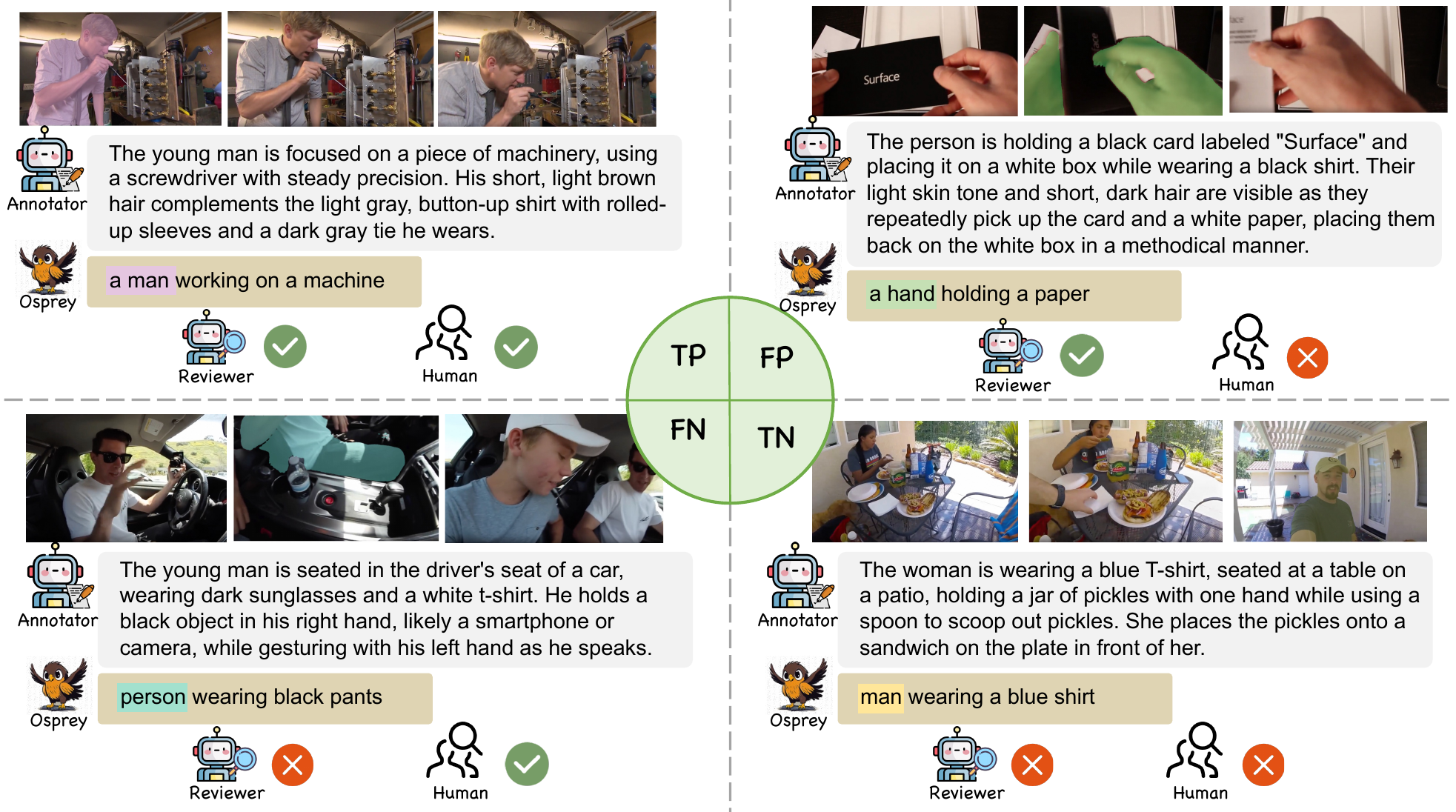}
\vspace{-1.0mm}
   \caption{Visual illustrations of human check process. TP, TN, FP and FN are introduced for the assessment on Reviewer.}
   \label{fig:human}
   \vspace{-1.8mm}
\end{figure*}

\textbf{Stage 1: Image-Text Alignment Pre-training.} In this initial pre-training phase, we utilize the same dataset as employed in the first stage of VideoLLaMA2.1~\cite{cheng2024videollama}. During this phase, the parameters of both the vision encoder and the large language model are frozen, and training is conducted solely on the STC connector~\cite{cheng2024videollama}, enabling the alignment of image and text modalities. 

\textbf{Stage 2: Region-Text Alignment Pre-training.} This stage further 
incorporates the Object Encoder to capture object-level features based on the weights obtained from Stage 1. The training focus is exclusively on the spatial-temporal  Object Encoder to ensure the alignment of intricate object-level features with corresponding language embeddings. We use the generated 500K region-level short descriptions, along with video and image referring segmentation datasets as the training data. During this stage, all the data are processed in single-frame mode to focus solely on alignment.

\textbf{Stage 2.5: High-Quality Knowledge Learning.} At this intermediate stage, the weights of vision encoder remain frozen, while the STC connector, Object Encoder, and LLM undergo fine-tuning. This stage aims to infuse the model with high-quality captioning data, utilizing a diverse dataset that includes 118K image-caption pairs, 30K video-caption pairs, 79K image-level region caption data, and 125K video-level region caption data, inclusive of the detailed descriptions we curated. For object-level video data, we employ a balanced approach, using half in single-frame mode and half in multi-frame mode.

\textbf{Stage 3: Visual Instruction Tuning.} 
The training configuration for this stage closely mirrors that of Stage 2.5. The primary objective is to enhance the model's ability to accurately interpret user instructions and tackle complex object-level understanding tasks. For video-level data, we utilize the same dataset segments as those used in VideoLLaMA2.1~\cite{cheng2024videollama}. For image-level data, we employ the datasets from LLaVA~\cite{liu2023improved}. In addition, we incorporate 294K image-level region data and 115K previously constructed video-level region data to further strengthen the model's capabilities. We also employ a balanced approach using half in single-frame mode and half in multi-frame mode in this stage.

\section{More Details of VideoRefer-700K Dataset and Benchmark}
\label{videoreferdataset}



\begin{figure*}[t]
  \centering
\includegraphics[width=0.999\linewidth]{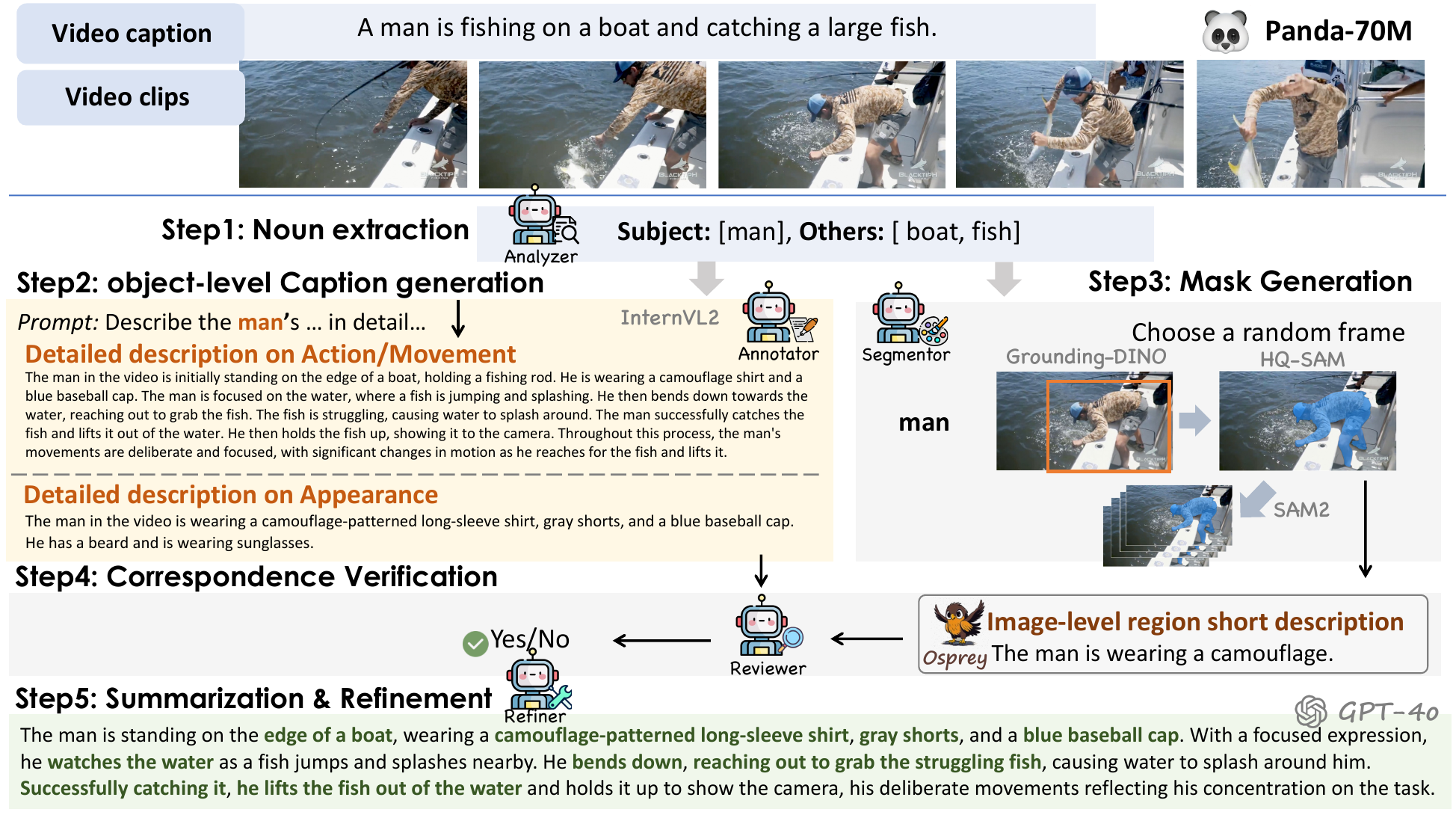}
   \caption{A detailed illustrative example of the construction pipeline in  our multi-agent data engine.}
   \label{fig:pipeline}
\end{figure*}

\subsection{Human Evaluation on Reviewer}
In our muliti-agent data engine, we introduce the Reviewer to address potential errors and mismatches, thereby ensuring the quality of our VideoRefer-700K dataset.
To assess the effectiveness of the Reviewer, we conducted a manual evaluation of its outputs.  We define the evaluation metrics as follows:

\begin{itemize}
  \item TP (True Positives): Items that the Reviewer identified as relevant and accurate, which are confirmed to be true upon manual inspection.
  \item TN (True Negatives): Items that the Reviewer discarded as irrelevant or inaccurate, which are indeed false according to the manual check.
  \item FP (False Positives): Items that the Reviewer considered as true, but are found to be false during manual verification.
  \item FN (False Negatives): Items that the Reviewer discarded as false, but are actually true upon manual review.

\end{itemize}

We randomly sampled 100 items each from both the data discarded and retained by the Reviewer. The detailed results are represented in Table~\ref{tab:confusion-matrix}, and the corresponding metrics are calculated as follows:

\begin{table}[t]
    \centering
    \begin{tabular}{l|cc}
        \toprule
        & \textbf{Manually True} & \textbf{Manually False} \\
        \hline
        \textbf{Reviewer True} & \cellcolor{green!10}{88 (TP)} &\cellcolor{red!10}{12 (FP)} \\
        \textbf{Reviewer False} & \cellcolor{red!10}{36 (FN)} & \cellcolor{green!10}{64 (TN)} \\
        \bottomrule
    \end{tabular}
    \caption{Confusion matrix of the randomly sampled 100 items in the Reviewer evaluation.}
    \vspace{-1.8mm}
    \label{tab:confusion-matrix}
\end{table}

\begin{equation}
\text{Accuracy} = \frac{TP + TN}{TP + TN + FP + FN} = 0.76,
\end{equation}
\begin{equation}
\text{Precision} = \frac{TP}{TP + FP} = 0.88,
\end{equation}

\begin{equation}
\text{Recall} = \frac{TP}{TP + FN} = 0.71,
\end{equation}

\begin{equation}
\text{F1 Score} = 2 \times \frac{\text{Precision} \times \text{Recall}}{\text{Precision} + \text{Recall}} = 0.79.
\end{equation}

The precision value stands at 88\%, indicating that the majority of samples identified as positive by the reviewer are indeed positive, thereby ensuring the data's quality.


\subsection{Example Illustrations}
\begin{figure*}[t]
  \centering
\includegraphics[width=0.999\linewidth]{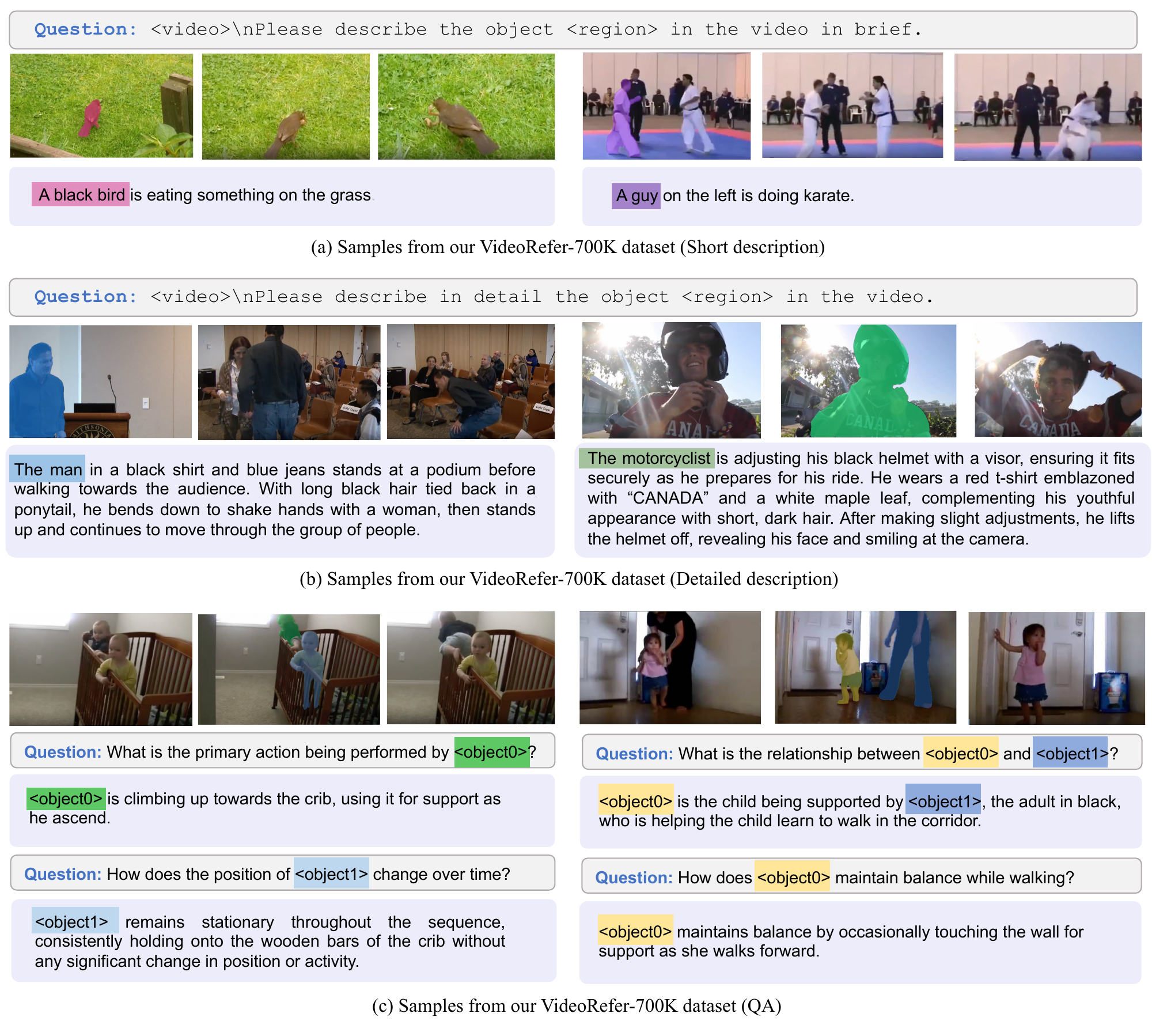}
     \vspace{-2.0mm}
   \caption{Visual samples from our VideoRefer-700 dataset, typical including short descriptions, detailed descriptions, and QA pairs.}
   \label{fig:dataset}
\end{figure*}

We provide a typical example to better exhibit the construction pipeline of our multi-agent data engine, as shown in Fig.~\ref{fig:pipeline}.
Additionally, the data distribution of our VideoRefer-700K dataset is illustrated in Fig.~\ref{fig:videorefer_data}. 
Fig.~\ref{fig:dataset} further showcases the additional visual samples from the VideoRefer-700K dataset.






\begin{figure*}[t]
  \centering
\includegraphics[width=0.999\linewidth]{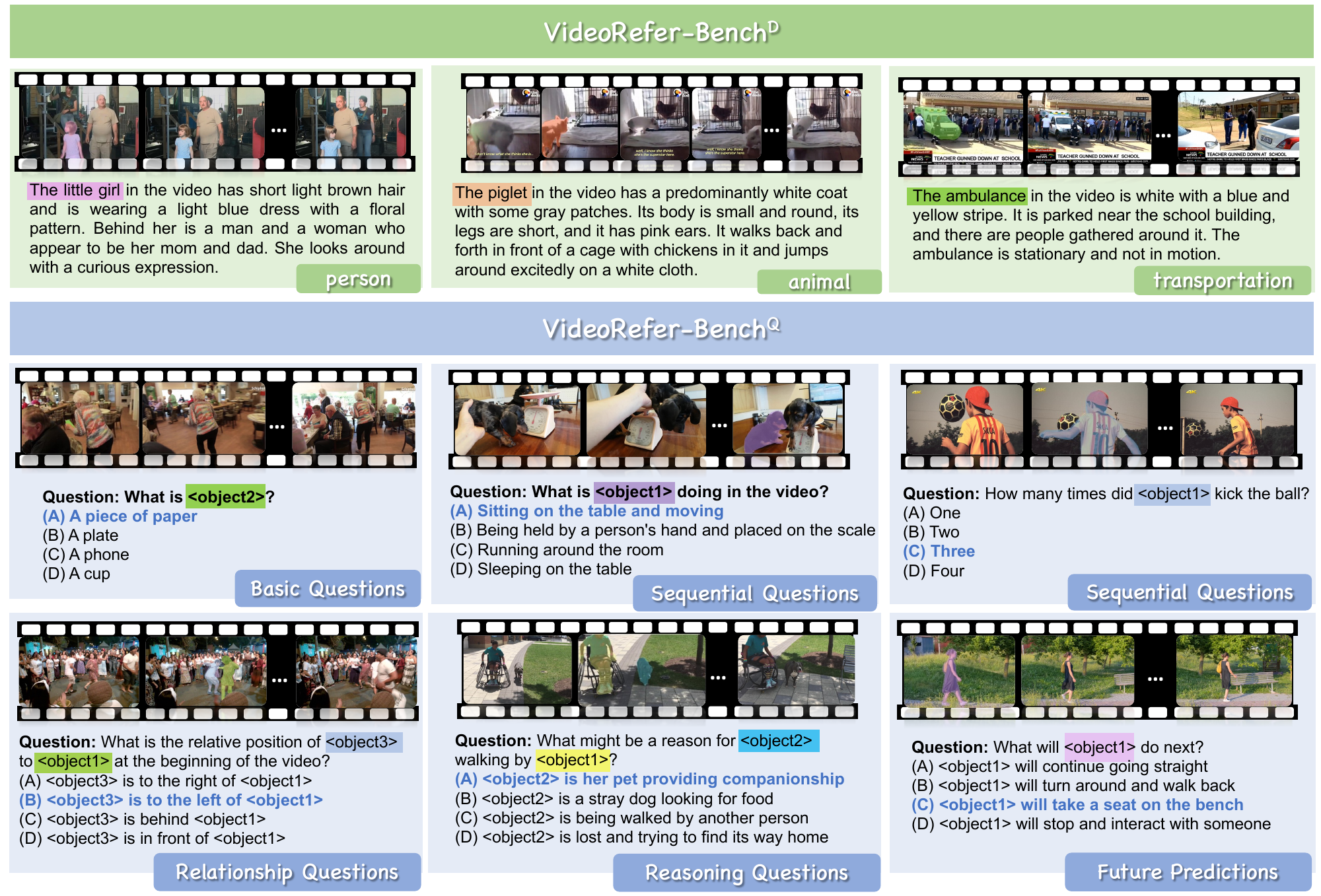}
\vspace{-1.0mm}
   \caption{Visual examples of our VideoRefer-Bench, including  VideoRefer-Bench$^\text{D}$ and VideoRefer-Bench$^\text{Q}$.}
   \label{fig:bench_vis}
   \vspace{-1.0mm}
\end{figure*}

\subsection{More Benchmark Visualization}
\label{benchmark}
We present more visualizations of our benchmark, VideoRefer-Bench$^\text{D}$ and VideoRefer-Bench$^\text{Q}$, as shown in Fig.~\ref{fig:bench_vis}.
These visualizations aim to provide a deeper understanding of benchmarks' structure and content. 
VideoRefer-Bench$^\text{D}$ focuses on detailed description tasks, facilitating the analysis of nuanced object references and relationships within videos. Meanwhile, VideoRefer-Bench$^\text{Q}$ is designed for question-and-answer scenarios, capturing the essence of interactive video comprehension.

\section{Limitations}
\label{limitation}
In this work, our VideoRefer is designed on object-level spatial-temporal video understanding, without the abilities on grounding. 
This limitation may affect the applicability of our method in real-world scenarios, which requires identifying and associating objects within their dynamic contexts. In the future work, we will address this gap by integrating grounding abilities into our framework, extending our dataset and benchmark to improve the system's overall utility in practical applications.

\end{document}